%% file: MLST-103133-R1.tex
\documentclass{article}

\PassOptionsToPackage{numbers, compress}{natbib}

\usepackage[preprint]{neurips_2024}

\usepackage[utf8]{inputenc} 
\usepackage[T1]{fontenc}    
\usepackage{hyperref}       
\usepackage{url}            

\usepackage{booktabs}       
\usepackage{amsfonts}       
\usepackage{nicefrac}       
\usepackage{microtype}      
\usepackage{xcolor}         

\usepackage{graphicx}
\usepackage{bm}
\usepackage{enumitem}
\usepackage{notoccite}
\usepackage{tabularx}

\usepackage{subfigure}

\usepackage{multirow}
\usepackage{multicol}

\usepackage{amsmath}
\usepackage{amssymb}
\usepackage{mathtools}
\usepackage{amsthm}
\usepackage{bm}
\usepackage{algorithm}
\usepackage{algcompatible}

\usepackage[many]{tcolorbox}    	

\usepackage{wrapfig}
\usepackage{adjustbox}
\usepackage{ulem, lipsum}
\usepackage{soulpos}

\newtcolorbox{boxA}{
    boxrule = 1.5pt,
    colframe = black 
}

\title{Stochastic Resetting Mitigates \\ Latent Gradient Bias of SGD from Label Noise}

\author{
  Youngkyoung Bae\textsuperscript{1}\footnotemark[1]\quad
  Yeongwoo Song\textsuperscript{2}\thanks{Equal Contribution.}\quad Hawoong Jeong\textsuperscript{2, 3}\thanks{Correspondence to Hawoong Jeong.}\\ 
  \textsuperscript{1}Department of Physics and Astronomy, Seoul National University \\ \quad 
  \textsuperscript{2}Department of Physics, KAIST \quad 
  \textsuperscript{3}Center for Complex Systems, KAIST\\
  \texttt{tardis\_95@snu.ac.kr, ywsong1025@kaist.ac.kr, hjeong@kaist.edu}\\
}

\begin{document}

\newcommand{\todo}[1]{\textcolor{red}{#1}}
\newcommand{\YB}[1]{{\textcolor{blue}{#1}}}
\newcommand{\ys}[1]{{\textcolor{green}{#1}}}
\newcommand{\redst}[1]{\setstcolor{red}\st{#1}}

\maketitle

\begin{abstract}
    Giving up and starting over may seem wasteful in many situations such as searching for a target or training deep neural networks (DNNs). 
    Our study, though, demonstrates that resetting from a checkpoint can significantly improve generalization performance when training DNNs with noisy labels.
    In the presence of noisy labels, DNNs initially learn the general patterns of the data but then gradually memorize the corrupted data, leading to overfitting.
    By deconstructing the dynamics of stochastic gradient descent (SGD), we identify the behavior of a latent gradient bias induced by noisy labels, which harms generalization.
    To mitigate this negative effect, we apply the stochastic resetting method to SGD, inspired by recent developments in the field of statistical physics achieving efficient target searches. We first theoretically identify the conditions where resetting becomes beneficial, and then we empirically validate our theory, confirming the significant improvements achieved by resetting.
    We further demonstrate that our method is both easy to implement and compatible with other methods for handling noisy labels.
    Additionally, this work offers insights into the learning dynamics of DNNs from an interpretability perspective, expanding the potential to analyze training methods through the lens of statistical physics.
\end{abstract}

\section{Introduction}
\label{sec:1}

When we explore a search space having complex choices of training schemes or search for appropriate hyperparameters of deep neural networks (DNNs), we often meet circumstances that cause us to give up and train the network all over again.
This is akin to our experiences in daily life, where we face various tasks that require solving problems through hit-and-miss. For example, when trying to find a beloved one's face in a crowd, our eyes typically flick back to a certain starting point after scanning the surrounding area. Similarly, when searching for a misplaced wallet after a big night out, we often fail to locate it and restart our search from some original location.
These patterns are also frequently observed in animal behavior, such as foraging for food and returning to familiar locations such as nests or dens.
In these situations, one might think that revisiting places is a waste of time and resources, potentially diminishing search performance.
However, recent developments in statistical physics have proven that resetting to the start or a mid-point can improve the performance of the search process, meaning that this strategy is not so haphazard after all.

This effect of resetting from a particular configuration has been extensively investigated in the field of statistical physics in recent years~\cite{evans2011diffusion, evans2020stochastic}.
These investigations typically involve a blind searcher who evolves their current state stochastically over time without knowledge of the target's location.
Surprisingly, it has been found that resetting does not hinder the search process but rather can make the searcher more efficient across diverse conditions, including scenarios with high dimensions or the presence of external forces~\cite{evans2011diffusionopt, evans2014diffusion, ray2019peclet, gupta2019stochastic, pal2020search, tal-friedman2020experimental, ray2020diffusion, bruyne2022optimal, negar2023stochastic}.
Capitalizing on the success of the resetting strategy, numerous algorithms incorporating this approach have begun to emerge in diverse fields such as molecular dynamics simulations~\cite{blumer2022stochastic, blumer2024combining} and queuing systems~\cite{bonomo2022mitigating}.

In parallel, statistical physics has emerged as a valuable framework for understanding the nature of DNNs, offering insights that are both explainable and interpretable~\cite{carleo2019machine, bahri2020statistical}.
Several techniques rooted in statistical physics, such as spin-glass theories~\cite{baity2018comparing, ghio2024sampling} and analytical tools for stochastic systems~\cite{yang2023stochastic, biroli2024dynamical}, have been recently applied to advance the understanding of DNNs. Yet it remains unclear how statistical physics, in addition to its ability to analyze the learning process of DNNs, can be effectively utilized to enhance their performance, especially in practical scenarios including low-quality datasets.

\begin{figure}[!t]
    \centering
    \includegraphics[width=\linewidth]{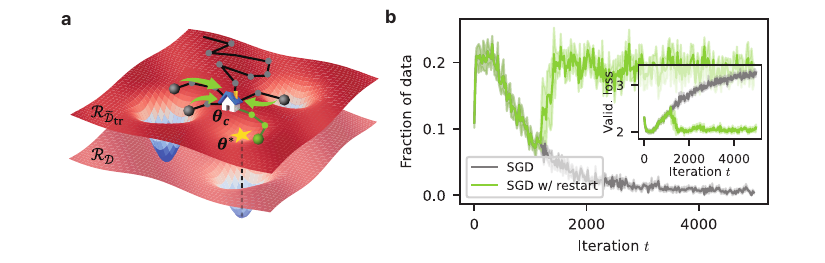}
    \vskip -0.1in
    \caption{
     (a) Schematic of stochastic gradient descent (SGD) dynamics with stochastic resetting. The network parameter vector $\bm{\theta}$ evolves via SGD to find an optimal value $\bm{\theta}^*$ on the training risk landscape $\mathcal{R}_{\tilde{\mathcal{D}}_{\rm tr}}$ (upper colormap), which differs from the true risk landscape $\mathcal{R}_{\mathcal{D}}$ (lower colormap) due to corrupted data.
     Here, $\bm{\theta}$ resets to the checkpoint $\bm{\theta}_c$ (home icon) with the reset probability $r$ and resets to $\bm{\theta}_c$.
    (b) Fraction of correctly predicted data with wrong labels during training with SGD (gray) and SGD with reset (green).
    The inset shows the validation losses during training.
    }\label{fig1}
    \vskip -0.1in
\end{figure}

In this work, we propose applying the stochastic resetting strategy to supervised learning with noisy labels and show that it can prevent overfitting to corrupted data (also called the memorization effect).
During network training, our method resets the model parameters to a checkpoint with a certain probability and restarts the training process [Fig.~\ref{fig1}(a)]. 
By mapping the stochastic gradient descent (SGD) dynamics to the corresponding Langevin dynamics, we explore in-depth to understand the mechanisms and conditions by which resetting can help SGD find the optimal parameters.
Our main contributions are summarized as follows.

\begin{itemize}
\item We reveal a latent gradient bias in the SGD dynamics induced by noisy labels, which drives the memorization effect in DNNs. Based on this finding, we apply the stochastic resetting method to counteract this effect and explain its core beneficial mechanism (Sec.~\ref{sec:3}).
\item We analyze the key factors for applying the stochastic resetting method. First, we discuss the means of selecting preferable checkpoints to reset to. Then, both theoretically and empirically, we find that the improvements of resetting increase as the stochasticity of the SGD dynamics and the proportion of corrupted training data increase (Sec.~\ref{sec:4.1}, \ref{sec:4.2}).
\item We show that the resetting method can be seamlessly integrated into existing approaches and consistently improves generalization performance across several standard benchmark datasets, including those incorporating real-world noise (Sec.~\ref{sec:4.4}).
\end{itemize}

\section{Related Works}
\label{sec:2}

\subsection{Search processes and stochastic resetting in statistical physics}

Search processes are ubiquitous across various domains, spanning from systems in nature to applications in engineering. 
For instance, ligands exhibit search processes as they navigate toward target binding sites within proteins~\cite{berg1981diffusion, coppey2004kinetics, ghosh2018first}, and similarly, predators employ search strategies to locate their prey in the wild~\cite{bartumeus2009optimal, viswanathan2011physics}. 
In engineering, search processes are relevant to finding primary research studies~\cite{dieste2009developing}, ranking web pages~\cite{brin1998anatomy}, and determining optimal hyperparameters for training algorithms~\cite{yu2020hyper}. 
Although diverse search strategies are employed depending on the problem at hand, they share a common goal: to identify an efficient search protocol. 
Efficiency is typically assessed by the time required to reach a target, referred to as the first passage time (FPT) in the context of random walk literature~\cite{redner2001aguide}.
Numerous search strategies have been investigated to achieve this goal, including the L\'evy strategies~\cite{viswanathan1999optimizing, lomholt2008levy}, self-avoiding walks~\cite{amit1983asymptotic, grassberger2017self}, intermittent strategies~\cite{benichou2011intermittent}, persistent random walks~\cite{tejedor2012optimizing}, and more~\cite{meyer2021optimal}.
One recent strategy that has garnered attention is stochastic resetting, with studies showcasing its ability to enhance search performance by revisiting previous places~\cite{evans2020stochastic, pal2020search, bae2022unexpected, bruyne2022optimal}. 
In particular, these studies have demonstrated that stochastic restarts prevent a random searcher from wandering too far, thereby ensuring a finite mean time to find a target, whereas the mean time is infinite for a diffusive particle without resetting.
Drawing from this concept, we introduce a resetting method for training DNNs and illustrate its effectiveness in addressing the noisy label problem.

\subsection{Deep learning from noisy labels}

While the accessibility of large datasets has propelled remarkable advancements in DNNs, the presence of noisy labels within these datasets often leads to erroneous model prediction~\cite{xiao2015learning}.
Specifically, DNNs tend to overfit the entire corrupted training dataset by memorizing the wrong labels, which degenerates their generalization performance on a test dataset.
Numerous studies have been conducted to address this overfitting phenomenon~\cite{han2020survey, hu2020simple, li2022selective, song2023learning, xia2023combating, huang2023twin}, and it has been revealed that DNNs initially learn the clean data (general patterns) during an early learning stage and then gradually memorize the corrupted data (task-specific patterns)~\cite{zhang2017understanding, arpit2017acloser, feng2021phases}.
The overfitting issue stemming from the memorization effect can be seen in Fig.~\ref{fig1}(b), where the model's accuracy in predicting the true labels of the data with noisy labels exhibits an inverted U-shaped curve as the model progressively memorizes the noise.
Based on this understanding, the surprising effectiveness of the early-stopping method~\cite{li2020gradient} in alleviating the memorization effect becomes evident; as such, various methods have been proposed to leverage this insight, including Co-teaching~\cite{han2018coteaching}, SELFIE~\cite{song2019selfie}, early learning regularization (ELR)~\cite{liu2020early-learning}, and robust early learning~\cite{xia2021robust}.
Our proposed method also capitalizes on this insight by enabling the DNN to reset to a checkpoint, i.e., previously visited parameters during early learning stages, and implicitly serves as a regularization mechanism by indirectly affecting the SGD dynamics.
Additionally, our theoretical analysis explores how label noise affects the performance of DNNs from the perspective of optimization strategies, such as those discussed in Refs.~\cite{li2020gradient, feng2021phases}. We adopt a different approach based on statistical physics, assuming more practical settings, which offers novel insights and leads to the development of our method.
In Sec.~\ref{sec:3}, we provide a detailed analysis identifying a latent gradient bias of SGD due to label noise that causes the memorization effect, and how stochastic resetting can mitigate such negative effects.

\section{Methodology}
\label{sec:3}
In this section, we first investigate the SGD dynamics in the presence of label noise and identify the latent gradient bias that leads to memorizing corrupted labels. We then introduce how stochastic resetting can be incorporated into SGD and demonstrate how this improves generalization performance by approximating SGD dynamics into Langevin dynamics.


\textbf{Problem setup.}\; Consider a $c$-class classification problem, which is a supervised learning task aimed at training a function to map input features to labels through a DNN.
Let $\mathcal{X} \subset \mathbb{R}^p$ be the feature space, $\mathcal{Y}=\{0, 1\}^{c}$ be the label space in one-hot vector form, and $\bm{f}_{\bm{\theta}}: \mathcal{X} \rightarrow \mathcal{Y}$ be a DNN model where $\bm{\theta}\in\mathbb{R}^{d}$ encompasses all trainable parameters in the DNN.
The goal is to find an optimal $\bm{\theta}^*$ such that $\bm{f}_{\bm{\theta}^*}$ accurately assigns labels to corresponding input features, given an unknown joint probability distribution $P_\mathcal{D}$ over $\mathcal{X} \times \mathcal{Y}$ [Fig.~\ref{fig1}(a)].
To obtain this, a training algorithm is applied to minimize the risk $\mathcal{R}_{\mathcal{D}} (\bm{\theta}) \equiv \left\langle \mathcal{L}\left( \bm{x}, \bm{y} ; \bm{\theta} \right) \right\rangle_{\mathcal{D}}$ during training, where $\mathcal{L}$ denotes a loss function (e.g., cross-entropy loss).
Here, $\mathcal{L}\left( \bm{x}, \bm{y} ; \bm{\theta} \right)$ denotes the loss for a sample $(\bm{x}, \bm{y})$ from $P_\mathcal{D}$ with a given model $\bm{f}_{\bm{\theta}}$, and $\langle \cdot \rangle_{\mathcal{D}}$ denotes the average over $P_\mathcal{D}$. 
In a typical classification problem, the DNN is trained by minimizing the risk on the training dataset $\mathcal{D}_{\rm tr}$ via SGD and $\bm{\theta}^*$ is selected at the minimum risk on the validation dataset $\mathcal{D}_{\rm val}$ to mitigate overfitting on $\mathcal{D}_{\rm tr}$, where $\mathcal{D}_{\rm tr (val)}=\{ (\bm{x}_i, \bm{y}_i) \}_{i=1}^{N_{\rm tr(val)}}$ and each $(\bm{x}_i, \bm{y}_i)$ is sampled from $P_{\mathcal{D}}$.
Empirically, the risk on the training (validation) dataset is computed as $\mathcal{R}_{\mathcal{D}_{\rm tr(val)}} (\bm{\theta}) = \left( 1/N_{\rm tr(val)} \right) \sum_{i=1}^{N_{\rm tr(val)}} \mathcal{L}\left( \bm{x}_i, \bm{y}_i ; \bm{\theta} \right)$. In the presence of noisy labels, suppose we have a corrupted training dataset $\tilde{\mathcal{D}}_{\rm tr}=\{ (\bm{x}_i, \tilde{\bm{y}}_i) \}_{i=1}^{N_{\rm tr}}$, where $\tilde{\bm{y}}$ is a noisy label that may be corrupted from a ground truth label $\bm{y}_i$, and $(\bm{x}_i, \tilde{\bm{y}}_i)$ is sampled from the corrupted distribution $P_{\tilde{\mathcal{D}}}$. 
This corrupted dataset can be partitioned into two subsets, i.e., $\tilde{\mathcal{D}}_{\rm tr} \equiv [\tilde{\mathcal{D}}_{\rm tr}^c, \tilde{\mathcal{D}}_{\rm tr}^w]$, where $\tilde{\mathcal{D}}_{\rm tr}^c$ ($\tilde{\mathcal{D}}_{\rm tr}^w$) consists of $N_{\rm tr}^c$ ($N_{\rm tr}^w$) samples with correct (wrong) labels.
Note that $N_{\rm tr}^c = (1-\tau) N_{\rm tr}$ and $N_{\rm tr}^w = \tau N_{\rm tr}$ for an unknown noise rate $\tau \in [ 0, 1 ]$.

\subsection{Latent gradient bias in SGD by label noise}
\label{sec:3.1}

When we apply the minibatch SGD to minimize the empirical risk $\mathcal{R}_{\mathcal{D}_{\rm tr}}(\bm{\theta})$ with respect to $\bm{\theta}$, the update rules of $\bm{\theta}$ at each training iteration $t$ can be represented by
\begin{equation}
\begin{aligned}
    \Delta \bm{\theta}_t &= -\frac{\eta}{B}\sum_{(\bm{x}_i, \bm{y}_i)\in\mathcal{B}_t} \bm{\nabla}_{\bm{\theta}} \mathcal{L}_i(\bm{\theta}_t) \\
    & = -\frac{\eta}{N_{\rm tr }}\sum_{(\bm{x}_i, \bm{y}_i)\in\mathcal{D}_{\rm tr}} \bm{\nabla}_{\bm{\theta}} \mathcal{L}_i(\bm{\theta}_t) + \left( \frac{\eta}{N_{\rm tr }}\sum_{(\bm{x}_i, \bm{y}_i)\in\mathcal{D}_{\rm tr}} \bm{\nabla}_{\bm{\theta}} \mathcal{L}_i(\bm{\theta}_t) -\frac{\eta}{B}\sum_{(\bm{x}_i, \bm{y}_i)\in\mathcal{B}_{t}} \bm{\nabla}_{\bm{\theta}} \mathcal{L}_i(\bm{\theta}_t) \right),
\end{aligned}
\label{eq:SGD}
\end{equation}
where $\bm{\theta}_t$ is $\bm{\theta}$ at the $t$-th iteration, $\Delta \bm{\theta}_t \equiv \bm{\theta}_{t+1}-\bm{\theta}_t$, and $\mathcal{L}_i(\bm{\theta}_t) \equiv \mathcal{L}(\bm{x}_i, \bm{y}_i ;\bm{\theta}_t)$ for simplicity. Here, $\eta > 0$ is the learning rate and $\mathcal{B}_t$ is the minibatch of size $B$ consisting of independent and identically distributed (i.i.d.) samples from $\mathcal{D}_{\rm tr}$.
While the first term on the right-hand-side (RHS) is deterministic for a given $\mathcal{D}_{\rm tr }$, the second term on the RHS is stochastic due to the randomly sampled batch at each iteration.
Thus, Eq.~\eqref{eq:SGD} can be rewritten as
\begin{equation}
\begin{aligned}
    \Delta \bm{\theta}_t = -\bm{\nabla}_{\bm{\theta}} \mathcal{R}_{\mathcal{D}_{\rm tr}}(\bm{\theta}_t) \eta + \bm{\xi}_t \sqrt{\eta},
\end{aligned}
\label{eq:SGD_LE}
\end{equation}
where a random noise vector $\bm{\xi}_t \equiv \sqrt{\eta} \left( \bm{\nabla}_{\bm{\theta}} \mathcal{R}_{\mathcal{D}_{\rm tr}}(\bm{\theta}_t) - \bm{\nabla}_{\bm{\theta}} \mathcal{R}_{\mathcal{B}_{t}}(\bm{\theta}_t) \right) \in \mathbb{R}^d$ satisfies $\langle \bm{\xi}_t \rangle_{\mathcal{D}_{\rm tr}} = \bm{0}$ and $\langle \bm{\xi}_t \bm{\xi}_s^{\rm T} \rangle_{\mathcal{D}_{\rm tr}} = 2 \mathsf{D}(\bm{\theta}_t) \delta_{ts}$ with $\mathsf{D}(\bm{\theta}_t) \equiv \eta \mathsf{\Sigma}(\bm{\theta}_t) /(2B) $, where $\delta_{ij}$ denotes the Kronecker delta (see details in the Supplementary Materials (SM) and Refs.~\cite{li2017stochastic, smith2018bayesian, ziyin2022strength}).
In terms of Langevin dynamics, $\mathcal{R}_{\mathcal{D}_{\rm tr}}(\bm{\theta})$ and $\mathsf{D}(\bm{\theta})$ correspond to the potential and diffusion matrix, respectively, where the former generates the deterministic long-term trend called drift and the latter determines the level of stochasticity of the system~\cite{gardiner2004handbook}. 
As a result, the SGD dynamics of $\bm{\theta}_t$ can be understood by the Langevin dynamics of a $d$-dimensional particle diffusing with drift $-\bm{\nabla}_{\bm{\theta}} \mathcal{R}_{\mathcal{D}_{\rm tr}}(\bm{\theta}_t)$ and diffusion matrix $\mathsf{D}(\bm{\theta}_t)$. 

\begin{figure}[!t]
    \centering
    \includegraphics[width=\linewidth]{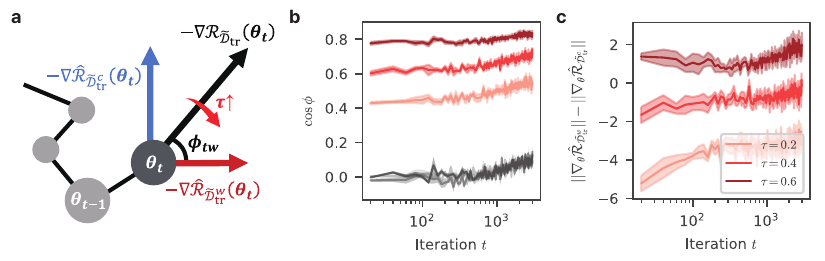}
    \vskip -0.1in
    \caption{
    (a) Schematic of $-\bm{\nabla}_{\bm{\theta}} \mathcal{R}_{\tilde{\mathcal{D}}_{\rm tr}}(\bm{\theta})$, decomposed by two orthogonal terms $-\bm{\nabla}_{\bm{\theta}}\hat{\mathcal{R}}_{\tilde{\mathcal{D}}_{\rm tr}^c}(\bm{\theta})$ and $-\bm{\nabla}_{\bm{\theta}}\hat{\mathcal{R}}_{\tilde{\mathcal{D}}_{\rm tr}^w}(\bm{\theta})$.
    (b) Cosine similarity between $-\bm{\nabla}_{\bm{\theta}} \mathcal{R}_{\tilde{\mathcal{D}}_{\rm tr}}(\bm{\theta})$ and $-\bm{\nabla}_{\bm{\theta}}\hat{\mathcal{R}}_{\tilde{\mathcal{D}}_{\rm tr}^w}(\bm{\theta})$ ($\cos{\phi_{tw}}$; red),
    and between $-\bm{\nabla}_{\bm{\theta}} \mathcal{R}_{\tilde{\mathcal{D}}_{\rm tr}^c}(\bm{\theta})$ and $-\bm{\nabla}_{\bm{\theta}}\hat{\mathcal{R}}_{\tilde{\mathcal{D}}_{\rm tr}^w}(\bm{\theta})$ (grey), throughout all training iterations for varying noise rate $\tau$.  
    (c) Magnitude difference between the two vectors $\| \bm{\nabla}_{\bm{\theta}}\hat{\mathcal{R}}_{\tilde{\mathcal{D}}_{\rm tr}^w}(\bm{\theta}) \| -\| \bm{\nabla}_{\bm{\theta}}\hat{\mathcal{R}}_{\tilde{\mathcal{D}}_{\rm tr}^c}(\bm{\theta}) \|$ throughout all training iterations for varying $\tau$. Here, we set the batch size to $B=8$ in Setting $1$ described in Sec.~\ref{sec:4}. Darker colors represent larger values of $\tau$ in (b, c). 
    }\label{fig2}
    \vskip -0.1in
\end{figure}

For a corrupted dataset $\tilde{\mathcal{D}}_{\rm tr}$, the equation of SGD dynamics remains analogous to Eq.~\eqref{eq:SGD_LE} when we substitute ${\mathcal{D}}_{\rm tr}$ with $\tilde{\mathcal{D}}_{\rm tr}$.
Then we can divide the drift vector $-\bm{\nabla}_{\bm{\theta}} \mathcal{R}_{\tilde{\mathcal{D}}_{\rm tr}}(\bm{\theta}_t)$ into two components, one originating from $\tilde{\mathcal{D}}_{\rm tr}^c$ and the other from $\tilde{\mathcal{D}}_{\rm tr}^w$ as follows [Fig.~\ref{fig2}(a)]:
\begin{equation}
\begin{aligned}
    \Delta \bm{\theta}_t = -\left[ \bm{\nabla}_{\bm{\theta}} \hat{\mathcal{R}}_{\tilde{\mathcal{D}}_{\rm tr}^c}(\bm{\theta}_t) + \bm{\nabla}_{\bm{\theta}} \hat{\mathcal{R}}_{\tilde{\mathcal{D}}_{\rm tr}^w}(\bm{\theta}_t) \right] \eta + \bm{\xi}_t \sqrt{\eta},
\end{aligned}
\label{eq:SGD_LE2}
\end{equation}
with the gradients from the correct part $\bm{\nabla}_{\bm{\theta}} \hat{\mathcal{R}}_{\tilde{\mathcal{D}}_{\rm tr}^c}(\bm{\theta}_t) \equiv (1-\tau)\bm{\nabla}_{\bm{\theta}} {\mathcal{R}}_{\tilde{\mathcal{D}}_{\rm tr}^c}(\bm{\theta}_t)$ and the gradients from the wrong part $\bm{\nabla}_{\bm{\theta}} \hat{\mathcal{R}}_{\tilde{\mathcal{D}}_{\rm tr}^w}(\bm{\theta}_t) \equiv \tau\bm{\nabla}_{\bm{\theta}} {\mathcal{R}}_{\tilde{\mathcal{D}}_{\rm tr}^w}(\bm{\theta}_t)$. We refer to the drift from the wrong part $-\bm{\nabla}_{\bm{\theta}} \hat{\mathcal{R}}_{\tilde{\mathcal{D}}_{\rm tr}^w}(\bm{\theta}_t)$ as the \textit{latent gradient bias} by label noise.
Note that $\langle \bm{\nabla}_{\bm{\theta}}{\mathcal{R}}_{\tilde{\mathcal{D}}_{\rm tr}^c}(\bm{\theta}_t) \rangle_{\mathcal{D}} = \langle \bm{\nabla}_{\bm{\theta}} {\mathcal{R}}_{{\mathcal{D}}_{\rm tr}}(\bm{\theta}_t)  \rangle_{\mathcal{D}}$, implying that $-\bm{\nabla}_{\bm{\theta}}\hat{\mathcal{R}}_{\tilde{\mathcal{D}}_{\rm tr}^c}(\bm{\theta}_t)$ reflects the gradients toward the true optimum, while $-\bm{\nabla}_{\bm{\theta}}\hat{\mathcal{R}}_{\tilde{\mathcal{D}}_{\rm tr}^w}(\bm{\theta}_t)$ reflects the gradients toward the false optimum by memorizing the noisy labels.
Additionally, we observe that $-\bm{\nabla}_{\bm{\theta}}\hat{\mathcal{R}}_{\tilde{\mathcal{D}}_{\rm tr}^c}(\bm{\theta}_t)$ and $-\bm{\nabla}_{\bm{\theta}}\hat{\mathcal{R}}_{\tilde{\mathcal{D}}_{\rm tr}^w}(\bm{\theta}_t)$ are orthogonal to each other [Fig.~\ref{fig2}(b) and Fig.~S.1 in the SM], leading to $-\bm{\nabla}_{\bm{\theta}} \mathcal{R}_{\tilde{\mathcal{D}}_{\rm tr}}(\bm{\theta}_t)$ being represented by the sum of two orthogonal vectors.
Thus, $-\bm{\nabla}_{\bm{\theta}} \mathcal{R}_{\tilde{\mathcal{D}}_{\rm tr}}(\bm{\theta}_t)$ becomes more correlated with $-\bm{\nabla}_{\bm{\theta}}\hat{\mathcal{R}}_{\tilde{\mathcal{D}}_{\rm tr}^w}(\bm{\theta}_t)$ as the noise rate $\tau$ increases.
Figure~\ref{fig2}(b) and (c) illustrate that $-\bm{\nabla}_{\bm{\theta}}\hat{\mathcal{R}}_{\tilde{\mathcal{D}}_{\rm tr}^w}(\bm{\theta}_t)$ becomes increasingly dominant so that the drift gradually tilts toward wrong directions as $\tau$ increases, where $\cos \phi_{tw}$ denotes the cosine similarity between $-\bm{\nabla}_{\bm{\theta}}\hat{\mathcal{R}}_{\tilde{\mathcal{D}}_{\rm tr}^c}(\bm{\theta}_t)$ and $-\bm{\nabla}_{\bm{\theta}}\hat{\mathcal{R}}_{\tilde{\mathcal{D}}_{\rm tr}^w}(\bm{\theta}_t)$, and $\| \cdot \|$ denotes the Euclidean norm of a vector.
This gradually tilting trend toward a wrong direction can also be observed with respect to iteration $t$, implying that $-\bm{\nabla}_{\bm{\theta}}\hat{\mathcal{R}}_{\tilde{\mathcal{D}}_{\rm tr}^w}(\bm{\theta}_t)$ becomes increasingly dominant as the learning process progresses beyond an early learning phase~\cite{liu2020early-learning}.
Therefore, in the presence of noisy labels, we can see that the latent gradient bias emerges and hinders the search for the optimal parameters $\bm{\theta}^*$.

We note that a similar analysis of the SGD dynamics using statistical physics was previously performed in Ref.~\cite{feng2021phases}. 
While both studies observed an increasing trend in the effect of latent gradient bias during the learning process, their findings on the cosine similarity between drift components, $\cos \phi_{cw}$, differ from ours.
To address this discrepancy, we conducted additional experiments and identified the contributing factors, as detailed in Sec.~D.2 of the SM.

\subsection{Stochastic resetting method}
\label{sec:3.2}

We now describe how the stochastic resetting method can be integrated into SGD. Based on this, we establish the specific premises of this work. 
Let $\bm{\theta}_c$ be a reset checkpoint and $r$ be the reset probability at each iteration $t$, where a checkpoint refers to previously visited model parameters during training.
By incorporating the resetting method, Eq.~\eqref{eq:SGD_LE} for $\tilde{\mathcal{D}}_{\rm tr}$ can be expressed as
\begin{equation}
\begin{aligned}
    \bm{\theta}_{t+1} = 
    \begin{cases}
        \bm{\theta}_c, &\text{with probability $r$}, \\
        \bm{\theta}_{t} -\bm{\nabla}_{\bm{\theta}} \mathcal{R}_{\tilde{\mathcal{D}}_{\rm tr}}(\bm{\theta}_t) \eta + \bm{\xi}_t \sqrt{\eta},
        &\text{otherwise, Eq.~\eqref{eq:SGD_LE2}}.
    \end{cases}
\end{aligned}
\label{eq:SGD_RestartLE}
\end{equation}
Below we provide the pseudo-code for SGD with stochastic resetting, Algorithm~\ref{alg:Sto-Re}.
Our implementation is publicly available at {\url{https://github.com/qodudrud/stochastic-resetting}}.

\begin{algorithm}[!b]
    \begin{algorithmic}[1]
    \REQUIRE{Corrupted training set $\tilde{\mathcal{D}}_{\rm tr}$, validation set $\mathcal{D}_{\rm val}$}, reset probability $r$, threshold $\mathcal{T}$.
    \State Initialize $\bm{\theta}_0$ and set $t = 0$, $\bm{\theta}_c = \text{None}$, and $\bm{\theta}_{\rm best} = \text{None}$
    \FOR{$t=0$ to $T$}
        \State Update $\bm{\theta}_{t} \leftarrow \bm{\theta}_t -\frac{\eta}{B}\sum_{(\bm{x}_i, \bm{y}_i)\in\mathcal{B}_t} \bm{\nabla}_{\bm{\theta}} \mathcal{L}_i(\bm{\theta}_t)$ where $\mathcal{B}_t$ is a randomly sampled batch from $\tilde{\mathcal{D}}_{\rm tr}$
        \IF{$\bm{\theta}_c \neq \text{None}$ and $rand(0, 1) < r$}
            \State Restart $\bm{\theta}_{t} \leftarrow \bm{\theta}_c$
        \ENDIF
        \State $\bm{\theta}_{\rm best} \leftarrow \text{Valid}(\bm{\theta}_{t}, \mathcal{D}_{\rm val})$ where $\text{Valid}(\bm{\theta}_{t}, \mathcal{D}_{\rm val})$ checks whether $\mathcal{R}_{\mathcal{D}_{\rm val}} (\bm{\theta}_t)$ is the minimum.
        \IF{$\bm{\theta}_{\rm best}$ remains unchanged for $\mathcal{T}$ iterations \textbf{or} 
 $\bm{\theta}_t = \bm{\theta}_{\rm best}$}
            \State Set the checkpoint $\bm{\theta}_c \leftarrow \bm{\theta}_{\rm best}$
        \ENDIF
    \ENDFOR
    \end{algorithmic}
\caption{Stochastic resetting}
\label{alg:Sto-Re}
\end{algorithm}

The SGD dynamics with stochastic resetting involves two processes: resetting from a checkpoint $\bm{\theta}_c$ with probability $r$ [top of Eq.~\eqref{eq:SGD_RestartLE}], and maintaining the SGD dynamics with probability $1-r$ [bottom of Eq.~\eqref{eq:SGD_RestartLE}]. 
Note that Eq.~\eqref{eq:SGD_RestartLE} shares the same form as the (overdamped) Langevin equation with Poissonian reset~\cite{evans2011diffusionopt}, and also that training DNNs to find optimal parameters can be likened to a search process for an unknown target.
These parallels imply that similar advantages of stochastic resetting may arise in the training process of DNNs as in the random search process of Langevin dynamics. To examine this hypothesis, we first explore the beneficial mechanism of stochastic resetting in the search process with a simplified case.

\begin{figure}[!t]
    \centering
    \includegraphics[width=\linewidth]{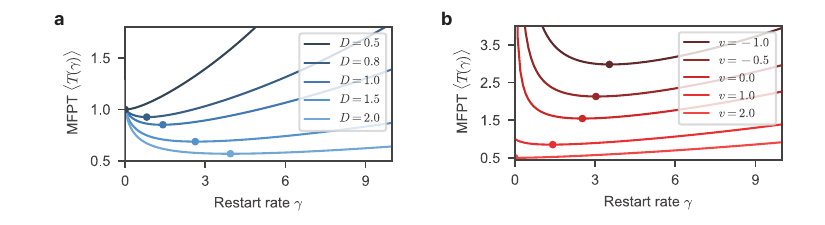}
    \vskip -0.1in
    \caption{ The mean first passage time (MFPT) $\langle T(\gamma) \rangle$ from Eq.~\eqref{eq:MFPT_drift} with varying (a) diffusion coefficient $D$ and (b) drift $v$ with respect to the reset rate $\gamma$. Markers represent the minimum MFPT, $\langle T(\gamma^*)\rangle$, at the optimal reset rate $\gamma^*$. We set $v=1$ in (a), $D=1$ in (b), and $L=1$ in both.
    }\label{fig3}
    \vskip -0.1in
\end{figure}

The search efficiency of a random search process is typically quantified by the mean first passage time (MFPT)~\cite{redner2001aguide}, which represents the average time to find a target. It has been well-established that incorporating stochastic resetting can significantly reduce the MFPT in various complex scenarios, including high-dimensional spaces~\cite{evans2011diffusion}, various confining potentials~\cite{pal2015diffusion, pal2020search}, and a searcher with momentum~\cite{gupta2019stochastic, singh2020random}, among others~\cite{evans2020stochastic}.
For a simplified case, let us consider a random searcher in one dimension with diffusion coefficient $D$, drift $v$, and reset rate $\gamma$ ($\equiv r/\Delta t$ with a time interval $\Delta t$ between steps), and assume that the searcher starts at the origin (reset point).
Then the MFPT for a target located at $L$ ($>0$) can be expressed by
\begin{equation}
\begin{aligned}
    \langle T(\gamma) \rangle = \frac{1}{\gamma} \left[ e^{ \frac{L}{2D} \left(\sqrt{v^2 + 4D\gamma} - v \right)}  - 1\right],
\end{aligned}
\label{eq:MFPT_drift}
\end{equation}
where $\langle T(\gamma) \rangle$ denotes the MFPT with the reset rate $\gamma$ (see the derivation in Sec.~B of the SM).
Examining Eq.~\eqref{eq:MFPT_drift} provides several insights into the effects of stochastic resetting.
When the random searcher either normally diffuses or drifts away from the target ($v \leq 0$ ), it is straightforward that $\langle T(\gamma) \rangle$ diverges as $\gamma \rightarrow 0$ but becomes finite for any $\gamma > 0$.
Conversely, when the searcher drifts toward the target ($v > 0$), while $\langle T(\gamma) \rangle$ is finite as $L/v$ without resetting, introducing stochastic resetting can significantly reduce $\langle T(\gamma) \rangle$ within a certain range of $\gamma$, provided the following condition is met:
\begin{equation}
\begin{aligned}
    {\rm Pe} \equiv \frac{L v}{2D} \leq 1.
\end{aligned}
\end{equation}
Here, ${\rm Pe}$ is known as the P\'eclet number, which quantifies the ratio between drift and diffusive transport rates, and the beneficial condition (${\rm Pe} \leq 1$) can be identified by verifying where $[d\langle T(\gamma) \rangle/d\gamma]|_{\gamma \rightarrow 0} < 0$.
Figure~\ref{fig3} illustrates how the behavior of $\langle T(\gamma) \rangle$ evolves with varying ${\rm Pe}$: as ${\rm Pe}$ decreases, the initially monotonically increasing curve gradually transforms into a U-shaped curve, achieving a minimum $\langle T(\gamma) \rangle$ at the optimal reset rate $\gamma^* > 0$.
These findings indicate that resetting is advantageous for a target search when the stochasticity ($D$) is sufficiently larger than the drift toward a target ($v$).
Furthermore, we note that both the optimal reset rate $\gamma^*$ and the improvement ratio $\langle T(0) \rangle/\langle T(\gamma^*) \rangle$ increase as ${\rm Pe}$ decreases (Fig.~\ref{fig3}), indicating that the benefits of resetting grow as $D$ increases and $v$ decreases.

According to the above observations, the beneficial properties of resetting in a random search process can be summarized as follows:
\begin{boxA}
When the stochasticity is sufficiently larger than the drift toward a target, resetting can be beneficial for random searches and there exists an optimal reset probability.
\end{boxA}
The advantageous mechanism of resetting in a random search process is to suppress trajectories that move away from the target, thus increasing the chances of finding it.
Drawing a parallel between the random search process and the training procedure of DNNs via SGD, we hypothesize this mechanism is also applicable to the noisy label problem in DNNs.
Specifically, in supervised learning with noisy labels, the stochasticity of the SGD dynamics increases as batch size $B$ decreases, and the drift component toward a target $-\bm{\nabla}_{\bm{\theta}} \hat{\mathcal{R}}_{\tilde{\mathcal{D}}^c_{\rm tr}}(\bm{\theta}_t)$ weakens (i.e., the latent gradient bias strengthens) as the noise rate $\tau$ increases.
Therefore, the boxed statement suggests that the resetting strategy would be beneficial for searching for optimal parameters in cases with a small batch size and large noise rate in the presence of noisy labels.
In Sec.~\ref{sec:4}, we perform several experiments and empirically show that stochastic resetting enhances generalization performance.

\section{Experiments}
\label{sec:4}
This section presents the experimental results that support our theory. We perform image classification tasks with noisy labels in the following settings.

\textbf{Setting 1} (Sec.~\ref{sec:4.1},~\ref{sec:4.2}, and \ref{sec:4.3}).
To examine the impact of stochastic resetting on the noisy label problem, we first utilize a small dataset called ciFAIR-10~\cite{barz2020do}, a variant of CIFAR-10~\cite{krizhevsky2009learning}.
We employ a vanilla convolutional neural network (VCNN, see Sec.~C.1 in the SM) to facilitate straightforward testing of our claims. 
Training is performed using cross-entropy loss, an SGD optimizer with a fixed learning rate of $10^{-2}$, and threshold $\mathcal{T}=1000$ iterations for the stochastic resetting method.
A clean validation set, $\mathcal{D}_{\rm val}$, is used to select the best model and monitor the validation loss during training.

\textbf{Setting 2} (Sec.~\ref{sec:4.4}).
We assess the generalization performance of our method on two benchmark datasets, CIFAR-10 and CIFAR-100~\cite{krizhevsky2009learning}, as well as its compatibility with various existing methods.
The model architecture used is ResNet-34~\cite{he2016deep}, trained with SGD using a momentum of $0.9$ and threshold $\mathcal{T}=5000$ iterations for the stochastic resetting method as default.
Additional details for the choice of hyperparameters are provided in Sec.~C.2 of the SM.
To demonstrate the efficacy of our method, we compare test accuracy with and without resetting.
Note that the optimizer and learning rate scheduler do not restart throughout this experiment.
To consider practical situations, a corrupted validation set, $\tilde{\mathcal{D}}_{\rm val}$, is used for model selection and validation loss monitoring during training.

\textbf{Setting 3} (Sec.~\ref{sec:4.5}). Under the same parameter conditions as Setting 2, we evaluate the performance of the stochastic resetting method on real-world noisy datasets, beyond the synthetic noise scenarios in Settings 1 and 2. Specifically, we use CIFAR-10N/100N, which are controllable, easy-to-use, and moderately sized real-world noisy datasets designed to enable fair comparisons across different benchmarks within accessible computational resources~\cite{wei2022learning}. 
These datasets contain real-world human annotation errors obtained from Amazon Mechanical Turk.
Additionally, we test on ANIMAL-10N, a dataset created by web-crawling image pairs of visually similar animals (e.g., cat and lynx, jaguar and cheetah)~\cite{song2019selfie}, to further examine its effectiveness across diverse real-world datasets.

In Settings 1 and 2, we apply symmetric noise with a noise rate $\tau$, where each label in $c$ classes is randomly flipped to an incorrect label in other classes with equal probability $\tau/(c-1)$.
Note that all results are obtained from the model at the optimal iteration based on minimum validation loss as default, and also that the resulting test accuracy is evaluated on the clean validation set, i.e., the test dataset $\mathcal{D}_{\rm te}$ is set to $\mathcal{D}_{\rm val}$.

We use the relative difference in validation loss (RDVLoss) and the relative difference in test accuracy (RDTAcc.) as metrics to indicate the relative improvement compared to the baseline. This metric enables effective comparison of the performance difference between stochastic resetting and original training.
These metrics are calculated by $[v(r)-v_{\rm base}]/v_{\rm base}$, where $v(r)$ is the resulting value with the reset probability $r$ and $v_{\rm base}$ is the baseline value obtained from the original training ($v_{\rm base} = v(0)$).
The unnormalized results can be found in Sec.~D.4 of the SM.
We repeated our experiments five times across all settings to report the average and standard error values.

\subsection{Which checkpoint would be preferable to reset to?}
\label{sec:4.1}

\begin{figure}[!t]
    \centering
    \includegraphics[width=\linewidth]{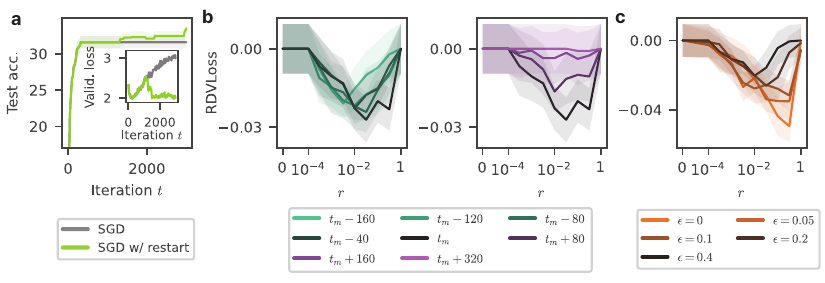}
    \vskip -0.1in
    \caption{(a) Test accuracies of the SGD (gray) and the SGD with our resetting method (green) during training. The inset shows the validation losses. (b,c) Relative difference of validation loss (RDVLoss) with varying the checkpoint to reset to with respect to the reset probability $r$.
    In (b), based on the checkpoint at the overfitting iteration $t_{m}$, RDVLoss is obtained in earlier iterations (left) and later iterations than $t_m$ (right).
    $t_m + \delta t$ denotes the iteration where the checkpoint is selected.
    In (c), RDVLoss is plotted with the perturbed checkpoint parameters $\bm{\theta}_{c, \epsilon} \equiv \bm{\theta}_c + \epsilon \hat{\bm{n}}$, where $\bm{\theta}_c$ denotes the checkpoint and $\hat{\bm{n}}$ denotes a random unit vector. The shaded areas denote the standard error.
    }\label{fig4}
    \vskip -0.1in
\end{figure}

To introduce the resetting strategy in DNN training, we first explore which checkpoint is suitable to reset to in order to find optimal parameters.
A straightforward choice is to select the parameters at the overfitting iteration $t_m$.
Here, the overfitting iteration $t_m$ refers to an iteration where the validation loss ceases to decrease and begins to increase due to the memorization effect [inset of Fig.~\ref{fig1}(b)].
The checkpoint at $t_m$, denoted by $\bm{\theta}_m$, has the minimum validation loss during training when the double descent phenomenon does not occur~\cite{nakkiran2020deep}, and is typically employed as an early-stopping point. 
Instead of early stopping and considering $\bm{\theta}_m$ as the final model, we utilize $\bm{\theta}_m$ as the checkpoint $\bm{\theta}_c$ to reset to (Algorithm~\ref{alg:Sto-Re}), leading to significantly improved results [Fig.~\ref{fig4}(a)].
Here, $\bm{\theta}_c$ is initially set to $\bm{\theta}_m$ and adaptively changes to the parameter at a newly found minimum validation loss during training.
As can be seen in Fig.~\ref{fig4}(a), resetting suppresses the trajectory of $\bm{\theta}$ to be near $\bm{\theta}_c$, which successfully prevents memorizing the noisy labels and increases the chance to find more appropriate parameters.
It is important to note that the reset probability $r$ controls the degree of suppression and the results at $r=0$ and $r=1$ are almost the same due to the overfitting phenomenon (the case of $r=1$ corresponds to the early-stopping method).
Therefore, the resulting RDVLoss curve for $r$ should be U-shaped, indicating that an optimal $r$ exists to optimize the performance.

While we simply select $\bm{\theta}_m$ as the initial $\bm{\theta}_c$ and adaptively update it, one may ask what effect the choice of $\bm{\theta}_c$ has.
To check this, we experiment with a fixed checkpoint both earlier and later than $\bm{\theta}_m$.
For earlier checkpoints [left panel in Fig.~\ref{fig4}(b)], the improvement over resetting from $\bm{\theta}_m$ slightly decreases and the value of the optimal $r$ gets smaller as the checkpoint gets earlier.
In contrast, for later checkpoints [right panel in Fig.~\ref{fig4}(b)], the improvement over resetting from $\bm{\theta}_m$ significantly decreases as the checkpoint gets later.
These results support our understanding of the beneficial mechanism of resetting in increasing the chance of finding better parameters, because the chance would decrease as the model memorizes more noise.
Therefore, we can conclude that resetting in early learning stages is a good choice: the more memorization occurs, the smaller the improvement.

We additionally experiment to verify how the effect of resetting changes with the distance between the (adaptive) checkpoint $\bm{\theta}_c$ at the minimum validation loss and a perturbed checkpoint $\bm{\theta}_{c, \epsilon}$. 
Here, we set the perturbed checkpoint by adding the perturbation $\epsilon \hat{\bm{n}}$ into $\bm{\theta}_c$ with varying the perturbation magnitude $\epsilon$, where $\hat{\bm{n}} \equiv \bm{n}/\|\bm{n}\| \in \mathbb{R}^d$ is a random unit vector with a standard normal random vector $\bm{n}$.
As shown in Fig.~\ref{fig4}(c), it is observed that the benefits of resetting decrease as the distance between the checkpoint to reset to and $\bm{\theta}_m$ increases.
This result also supports that while resetting can improve the generalization performance, the choice of checkpoint to reset to can affect the performance, and that the minimum validation loss point is a good choice.

\subsection{Impact of stochasticity and drift on stochastic resetting}
\label{sec:4.2}

As the statement highlighted in Sec.~\ref{sec:3.2} clarifies, it has been proven in the statistical physics field that the resetting strategy can improve search efficiency as the stochasticity becomes larger than the drift component toward a target.
In this section, we verify whether this statement is also valid in the training of DNNs and show under what circumstances resetting is more effective than not resetting.

\begin{figure}[!t]
    \centering
    \includegraphics[width=\linewidth]{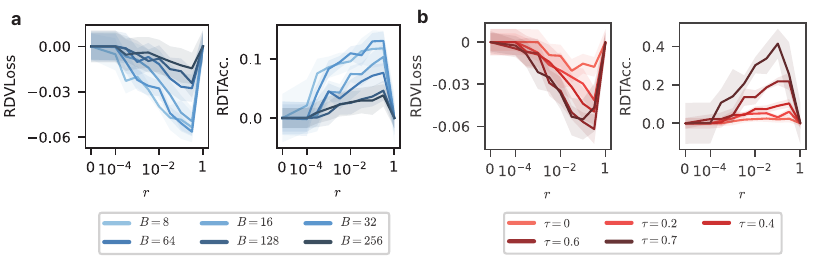}
    \vskip -0.1in
    \caption{Relative difference of validation loss (RDVLoss, left) and relative difference of test accuracy (RDTAcc., right) results with (a) varying the batch size $B$, and (b) varying the noise rate $\tau$ with respect to the reset probability $r$. We set $\tau = 0.4$ in (a) and $B=16$ in (b). The shaded areas denote the standard error.
    }\label{fig5}
    \vskip -0.1in
\end{figure}

It is first important to note that the stochasticity and the drift toward a target, i.e., $-\bm{\nabla}_{\bm{\theta}} \hat{\mathcal{R}}_{\tilde{\mathcal{D}}_{\rm tr}^c}$, can be controlled by the batch size $B$ and the noise rate $\tau$, respectively, as illustrated in Sec.~\ref{sec:3.1}.
Particularly, $\mathsf{D}(\bm{\theta}_t) \propto 1/B$ and $\bm{\nabla}_{\bm{\theta}} \hat{\mathcal{R}}_{\tilde{\mathcal{D}}_{\rm tr}^c}(\bm{\theta}_t) \propto 1-\tau$, meaning that the stochasticity increases and the drift toward a target decreases as $B$ decreases and $\tau$ increases, respectively. 
We quantitatively examine the RDVLoss and the RDTAcc. values with varying $B$ and $\tau$ with respect to the reset probability $r$.
Remarkably, the improvements of RDVLoss and RDTAcc. with resetting become more significant as $B$ decreases [Fig.~\ref{fig5}(a)] and $\tau$ increases [Fig.~\ref{fig5}(b)].
These observations strongly support our claim that stochastic resetting offers more benefits as the stochasticity increases and the drift toward a target decreases.

Moreover, we expect that the optimal reset probability $r^*$ decreases as $B$ decreases and $\tau$ increases, but this can only be verified qualitatively because the fluctuation of the results makes it difficult to identify $r^*$. 

\subsection{Ablation study on partial resetting}
\label{sec:4.3}

Until now, we have leveraged the memorization effect in our algorithm by utilizing the parameters at the minimum validation loss as the checkpoint for resetting the entire network, a process referred to as full resetting.
However, several studies have highlighted that different layers within a DNN exhibit varied learning behaviors, leading to distinct levels of overfitting across these layers~\cite{bai2021understanding, chen2023which}.
A prevailing explanation for this phenomenon suggests that gradients tend to weaken as they propagate from the latter layers (closer to the output layer) to the former layers (closer to the input layer).

\begin{wrapfigure}{t!}{0.5\textwidth} 
    \centering
    \includegraphics[width=0.5\textwidth]{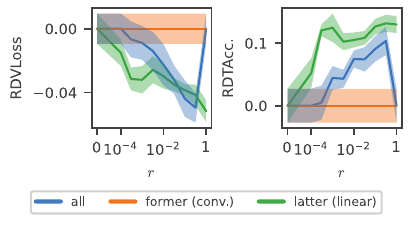}
    \vskip -.1in
    \caption{Relative difference of validation loss (RDVLoss) and relative difference of test accuracy (RDTAcc.) results with varying one section of the network to reset with respect to the reset probability $r$. We set $\tau = 0.4$ and $B=16$. The shaded areas denote the standard error.}
    \vskip -.2in
    \label{fig6}
\end{wrapfigure}

Here, we experiment on which layers, former or latter, play a more dominant role in improving performance with the resetting method.
For this, we introduce partial resetting, which involves resetting only one section of the network layers rather than the entire network, while the remaining section of layers continues to follow the standard SGD update rule without resetting.
We divide the VCNN structure into former and latter sections, comprising convolutional and linear layers, respectively, and apply partial resetting to one section.
Interestingly, our experiments reveal that partial resetting of the latter section can further enhance generalization performance compared to full resetting, whereas partial resetting of the former section does not yield improvements over the case with no resetting ($r=0$) (Fig.~\ref{fig6}).

Moreover, even when we freeze the latter section by setting
$r=1$, partial resetting of the latter section still achieves significant improvement.
We attribute these findings to a well-established observation: the former layers of CNNs tend to learn general features, while the latter layers tend to specialize in capturing specific features~\cite{yosinski2014how, cohen2018dnn, ansuini2019intrinsic, maennel2020what, bai2021understanding}.
In other words, the latter section composed of linear layers exhibits strong memorization of the corrupted data, leading to improved performances even when we freeze the latter section ($r=1$), whereas the former section composed of convolutional layers focuses on learning general features, leading to no improvements even with resetting.

It is essential to note that although our ablation study suggests the effectiveness of partial resetting, our findings do not imply that partial resetting of the latter section always enhances generalization performance compared to full resetting.
The extent to which each layer overfits the corrupted data depends on multiple factors, such as the network structure and the choice of loss function.
Thus, determining the most effective section of the network to reset also hinges on the specific context.
Future research investigating these points would be intriguing and valuable.

\subsection{Results on corrupted benchmark datasets}
\label{sec:4.4}

In Setting 2, we investigate the impact of the stochastic resetting strategy on the performance of benchmark datasets, CIFAR-10 and CIFAR-100, under symmetric noise corruption using various methods.
Table~\ref{table:FinalTable} compares the best results without resetting (No) and with resetting (Reset) at the reset probability $r=0.001$.
In the table, CE denotes cross-entropy loss, PartRestart denotes cross-entropy loss with partial resetting of only the last linear layer and the last two blocks of ResNet-34 (Sec.~\ref{sec:4.3}), MAE denotes robust mean absolute error~\cite{ghosh2017robust}, GCE denotes generalized cross-entropy~\cite{zhang2018generalized}, SL denotes symmetric cross-entropy loss~\cite{wang2019symmetric}, ELR denotes early-learning regularization~\cite{liu2020early-learning}, and SOP+ denotes sparse over-parameterization with consistency regularization and class-balance regularization~\cite{liu2022robust}.  
ELR and SOP+ are representative methods to robustly train DNNs with an additional regularization term to prevent overfitting to corrupted data. 
However, these methods require additional hyperparameters to be fine-tuned depending on the loss landscape (e.g., dataset, model architecture), and in practical scenarios, it is often costly to find the optimal settings of these methods. In this section, we consider both optimal and non-optimal hyperparameter settings to take such practical situations into account, where we use an asterisk (*) to denote a non-optimal hyperparameter setting (i.e., ELR*, SOP+*).
We provide additional details about the hyperparameters for each method in Sec.~C.2 of the SM.

\begin{table}[!t]
\caption{Test accuracies (\%) on test datasets with different methods. We compare the performance without resetting (No) and with resetting (Reset) at $r=0.001$. Results are presented as the average and the standard deviation. The best results are indicated in \textbf{bold} with statistical significance.
}
\centering
\resizebox{\columnwidth}{!}{
\begin{tabular}{cccccccc}
\hline
\rule{0pt}{2.0ex}
\multirow{2}{*}{Dataset} & \multirow{2}{*}{Method} & \multicolumn{2}{c}{Noise rate $\tau=0.2$} & \multicolumn{2}{c}{Noise rate $\tau=0.4$} & \multicolumn{2}{c}{Noise rate $\tau=0.6$}\\
& & No & Reset & No & Reset & No & Reset \\
\hline
\rule{0pt}{2.0ex}
    \multirow{7}{*}{CIFAR-10}   & CE & $84.8\pm0.4$ & $\bm{90.0\pm0.5 ^{\scriptscriptstyle ***}}$ & $80.8\pm0.7$ & $\bm{86.3\pm0.7 ^{\scriptscriptstyle ***}}$ & $72.9\pm0.7$ & $\bm{79.5\pm0.9^{\scriptscriptstyle ***}}$ \\
    & Part & --- & $\bm{88.5\pm0.4^{\scriptscriptstyle ***}}$ & --- & $\bm{83.7\pm0.9^{\scriptscriptstyle ***}}$ & --- & $\bm{75.1\pm0.4^{\scriptscriptstyle ***}}$ \\
    & MAE & $91.0\pm0.2$ & $91.0\pm0.2$ & $85.6\pm3.3$ & $\bm{85.9\pm3.3}$ & $67.1\pm6.8$ & $\bm{67.3\pm6.7}$ \\
    & GCE & $90.6\pm0.1$ & $\bm{91.0\pm0.2^{\scriptscriptstyle *}}$ & $85.3\pm0.4$ & $\bm{87.1\pm0.3^{\scriptscriptstyle ***}}$ & $76.2\pm0.5$ & $\bm{79.1\pm0.8^{\scriptscriptstyle ***}}$ \\
    & SL & $91.3\pm0.3$ & $\bm{91.4\pm0.1}$ & $86.6\pm0.2$ & $\bm{87.9\pm0.2^{\scriptscriptstyle ***}}$ & $80.1\pm0.5$ & $\bm{81.6\pm0.6^{\scriptscriptstyle **}}$ \\
    & ELR*  & $91.2\pm0.1$ & $\mathbf{91.3\pm0.2}$ & $88.8\pm0.2$ & $\mathbf{88.9\pm0.2}$ & $84.6\pm0.5$ & $84.6\pm0.5$ \\
    & SOP+* & ${94.1\pm0.2}$ & $94.1\pm0.2$ & $89.9\pm0.3$ & $89.9\pm0.3$ & $85.0\pm0.3$ & $\mathbf{85.1\pm0.3}$ \\
    \hline       
    \rule{0pt}{2.0ex}
    \multirow{7}{*}{CIFAR-100}  & CE & $62.7 \pm 5.6$ & $\bm{64.5 \pm 1.5}$ & $45.6 \pm 2.3$ & $\bm{56.9 \pm 3.0^{\scriptscriptstyle ***}}$ & $32.9 \pm 1.6$ & $\bm{44.1 \pm 3.1^{\scriptscriptstyle ***}}$ \\
    & Part & --- & $\bm{64.0\pm0.9}$ & --- & $\bm{55.5\pm2.1^{\scriptscriptstyle ***}}$ & --- & $\bm{43.5\pm1.5^{\scriptscriptstyle ***}}$ \\
    & MAE & $19.9 \pm 2.8$ & $19.9 \pm 2.8$ & $\bm{11.0 \pm 3.8}$ & $10.8 \pm 3.7$ & $6.7 \pm 1.2$ & $\bm{6.8 \pm 1.3}$ \\
    & GCE & $68.3 \pm 0.4$ & $\bm{69.3 \pm 0.2^{\scriptscriptstyle **}}$ & $61.2 \pm 0.6$ & $\bm{63.2 \pm 0.3^{\scriptscriptstyle ***}}$ & $50.1 \pm 0.6$ & $\bm{53.1 \pm 0.8^{\scriptscriptstyle ***}}$ \\
    & SL & $66.8\pm0.5$ & $\bm{68.6\pm0.6^{\scriptscriptstyle ***}}$ & $60.4\pm0.5$ & $\bm{63.1\pm0.7^{\scriptscriptstyle ***}}$ & $50.4\pm0.8$ & $\bm{54.1\pm0.8^{\scriptscriptstyle ***}}$ \\
    & ELR*  & $67.4\pm0.3$ & $\mathbf{70.4\pm0.3^{\scriptscriptstyle ***}}$ & $55.1\pm0.6$ & $\mathbf{64.1\pm0.6^{\scriptscriptstyle ***}}$ & $45.9\pm0.7$ & $\mathbf{54.0\pm1.0^{\scriptscriptstyle ***}}$ \\
    & SOP+* & ${72.1\pm0.3}$ & $72.1\pm0.3$ & $59.4\pm0.5$ & $\mathbf{65.7\pm0.9^{\scriptscriptstyle ***}}$ & $48.4\pm1.9$ & $\mathbf{54.8\pm2.3^{\scriptscriptstyle **}}$ \\
    \hline
\end{tabular}}
\label{table:FinalTable}
\end{table}

Remarkably, in all cases examined, our resetting method consistently achieves either at least equivalent or higher test accuracies compared to the baseline approach involving no resetting (Table~\ref{table:FinalTable}). 
Results show that the extent of improvement becomes more pronounced as the noise rate increases, which supports our claim that resetting becomes more advantageous with higher noise rates.
Furthermore, while the PartRestart method also obtains improved performance, it does not surpass the benefits of full resetting.
We conjecture that the network structure may influence the extent of the additional improvements obtained, as the memorization effect in different layers can vary depending on the network architecture~\cite{li2021how}.
Minimal improvements are observed in the MAE results for both datasets; this is primarily because the MAE convergence is too slow to identify a suitable checkpoint for resetting, consequently resulting in few instances of resetting in many trials.
For ELR and SOP+, no significant improvements are found in the performance when stochastic resetting is used with the known optimal hyperparameter settings, as shown in Table~S.4 in the SM. However, for ELR* and SOP+*, incorporating the stochastic resetting method leads to either at least equivalent performance or significant improvements compared to no resetting.
This can be explained by the design of ELR and SOP+ that provide regularization terms to prevent memorization, akin to reducing the effect of the latent gradient bias. These results indicate that stochastic resetting is compatible with well-established methods, and can also provide at least equal performance compared to the methods without resetting.

We likewise demonstrate the effectiveness of the stochastic resetting method in asymmetric (i.e., class-dependent) noise scenarios (Table~S.5 in the SM), and we also verify the improvement of the validation loss when using our stochastic resetting method (Table~S.6 in the SM).

\subsection{Results on real-world noisy datasets}
\label{sec:4.5}

\begin{table}[!t]
\caption{Test accuracies (\%) on test datasets with real-world datasets CIFAR-10N/100N. We compare the performance without resetting (No) and with resetting (Reset) at $r=0.001$. Results are presented as the average and the standard deviation. The best results are indicated in \textbf{bold} with statistical significance.
}
\centering
\resizebox{\columnwidth}{!}{
\begin{tabular}{ccccccccc}
\hline
\rule{0pt}{2.0ex}
\multirow{3}{*}{Method} & \multicolumn{2}{c}{CIFAR-10N} & \multicolumn{2}{c}{CIFAR-10N} & \multicolumn{2}{c}{CIFAR-10N} & \multicolumn{2}{c}{CIFAR-100N}\\
                        & \multicolumn{2}{c}{Random 1 ($\tau \sim 0.09$)} & \multicolumn{2}{c}{Aggregate ($\tau \sim 0.17$)} & \multicolumn{2}{c}{Worst ($\tau \sim 0.4$)} & \multicolumn{2}{c}{Noisy ($\tau \sim 0.4$)} \\
                        & No & Reset & No & Reset & No & Reset & No & Reset \\
\hline
\rule{0pt}{2.0ex}
                     CE & $82.8\pm0.5$ & $\bm{87.8\pm0.7^{\scriptscriptstyle ***}}$ & $86.1\pm0.6$ & $\bm{90.2\pm0.3^{\scriptscriptstyle ***}}$ & $73.8\pm1.9$ & $\bm{79.7\pm0.3^{\scriptscriptstyle ***}}$ & $46.7\pm1.6$ & $\bm{56.0\pm1.3^{\scriptscriptstyle ***}}$\\
                     MAE & $90.9\pm0.3$ & $90.9\pm0.2$ & $90.0\pm0.5$ & $90.0\pm0.5$ & $60.2\pm3.6$ & $60.2\pm3.6$ & $4.0\pm1.3$ & $4.0\pm1.3$\\
                     SL & $91.7\pm0.2$ & $\bm{91.8\pm0.2}$ & $89.6\pm0.2$ & $\bm{89.9\pm0.4}$ & $80.6\pm0.5$ & $\bm{81.3\pm0.8}$ & $52.3\pm0.5$ & $\bm{56.5\pm0.6^{\scriptscriptstyle ***}}$\\
                     GCE & $91.7\pm0.2$ & $\bm{91.9\pm0.2}$ & $89.4\pm0.4$ & $\bm{89.8\pm0.7}$ & $78.1\pm0.9$ & $\bm{82.3\pm0.5^{\scriptscriptstyle ***}}$ & $55.8\pm0.5$ & $\bm{58.0\pm1.2^{\scriptscriptstyle ***}}$\\
                     ELR* & ${89.5\pm0.3}$ & ${89.5\pm0.3}$ & ${91.3\pm0.2}$ & ${91.3\pm0.2}$ & ${81.9\pm0.4}$ & ${81.9\pm0.5}$ & $56.9\pm0.4$ & $\bm{60.5\pm0.5^{\scriptscriptstyle ***}}$\\
                     SOP+* & $91.2\pm0.2$ & ${91.2\pm0.2}$ & $\bm{93.2\pm0.1}$ & ${93.0\pm0.1}$ & $\bm{82.4\pm0.5}$ & ${82.3\pm0.7}$ & $57.0\pm0.4$ & $\bm{61.2\pm0.9^{\scriptscriptstyle ***}}$\\
                     ViT-T$_{\rm rd}$ & $63.2\pm0.9$ & $\bm{65.7\pm0.7^{\scriptscriptstyle ***}}$ & ${61.0\pm0.9}$ & $\bm{63.6\pm0.3^{\scriptscriptstyle ***}}$ & ${55.4\pm0.2}$ & $\bm{57.2\pm0.2^{\scriptscriptstyle **}}$ & $28.6\pm0.5$ & $\bm{32.8\pm0.5^{\scriptscriptstyle ***}}$\\
                     ViT-T$_{\rm pt}$ & $92.6\pm0.3$ & $\bm{93.6\pm0.6^{\scriptscriptstyle *}}$ & ${91.8\pm0.5}$ & $\bm{93.0\pm0.8^{\scriptscriptstyle **}}$ & ${85.9\pm1.4}$ & $\bm{87.4\pm0.7^{\scriptscriptstyle *}}$ & $63.8\pm1.1$ & $\bm{68.6\pm0.5^{\scriptscriptstyle **}}$\\
\hline
\end{tabular}}
\vskip -0.1in
\label{table:realworld}
\end{table}
\begin{wraptable}{r}{.5\textwidth}
\vskip -0.26in
\caption{Test accuracies (\%) on test datasets with real-world datasets ANIMAL-10N. We compare the performance without resetting (No) and with resetting (Reset) at $r=0.001$.}
\centering
\vskip 0.1in
\resizebox{.48\columnwidth}{!}
{
\begin{tabular}{ccc}
\hline
\rule{0pt}{2.0ex}
\multirow{1}{*}{Dataset} & No & Reset\\
\hline
\rule{0pt}{2.0ex}
    \multirow{1}{*}{ANIMAL-10N}  & $80.6\pm0.6$ & $\bm{85.1\pm0.4 ^{\scriptscriptstyle ***}}$\\
    \hline       
\end{tabular}}
\label{table:animal10n}
\end{wraptable}

Finally, we test our stochastic resetting method on real-world noisy datasets, namely CIFAR-10N/100N (Setting 3).
There are five different noise types for CIFAR-10N, namely Rand1, Rand2, Rand3, Aggregate, and Worst with a noise rate of 9.03\%, 17.23\%, 18.12\%, 17.64\%, and 40.21\%, respectively, and a single noise type for CIFAR-100N with a noise rate of 40.20\%. 
For CIFAR-10N, we selected Rand1, Aggregate, and Worst noise types to account for various real-world noise rates.
We further evaluate our method on ANIMAL-10N with the default setting of Setting 3 using cross-entropy loss.
For the ANIMAL-10N dataset, there is only a single noise type while the ground-truth labels remain unknown. The noise rate $\tau$ was estimated as $\tau \sim 0.08$ using cross-validation with a grid search and $\tau \sim 0.06$ based on human inspection, respectively.

Table~\ref{table:realworld} presents the performance results without resetting (No) and with resetting (Reset) at the reset probability $r=0.001$ for CIFAR-10N/100N.
Similarly, the results for ANIMAL-10N are shown in Table~\ref{table:animal10n}.
We provide additional details about the hyperparameters in Sec.~C.2 of the SM.
For the case of CIFAR-10N, the stochastic resetting method consistently achieves either at least equivalent or higher test accuracies compared to the baseline without resetting, similar to the results in Sec.~\ref{sec:4.4} (Table~\ref{table:FinalTable}). While the experiments on CIFAR-10N show some cases with minimal improvements from the stochastic resetting method, the results on CIFAR-100N and ANIMAL-10N provide strong evidence of the advantages of applying stochastic resetting in practical scenarios.
These results demonstrate that the stochastic resetting method can provide significant improvements while acting as a safeguard to maintain baseline performance.

Furthermore, to assess the applicability of our method beyond CNN-based approaches, we evaluate it on the Vision Transformer (ViT)~\cite{dosovitskiy2021an}.
As shown in Table 2, the resetting method consistently improves performance on ViT, denoted by ViT-T$_{{\rm rd}}$ and ViT-T$_{{\rm pt}}$, which correspond to ViT-Tiny models~\cite{rw2019timm} trained from randomly initialized and pre-trained weights, respectively.
Previous studies have reported that ViTs without pre-training are more prone to overfitting, particularly in the presence of label noise~\cite{chen2021an, khanal2024investigating}, which we also observe in Table~\ref{table:FinalTable}.
Interestingly, regardless of initialization, the resetting method results in significant performance improvements, highlighting its potential for a broader range of recent architectures and scenarios.

\section{Discussion}
\label{sec:5}

In this work, we identified a latent gradient bias that hinders SGD from generalizing in the presence of label noise. To address this, we developed a stochastic resetting method, motivated by the success of the resetting strategy in statistical physics.
By analyzing SGD dynamics through the lens of Langevin dynamics, we theoretically identified factors that influence the effectiveness of resetting, i.e., batch size and noise rate, and then experimentally confirmed the impact of these factors.
Experiments showed that the resetting method consistently yields equivalent or improved performance on benchmark datasets compared to existing methods.

As our method can be implemented with minimal code changes and without additional computational costs, flexibility and ease of integration with other approaches are ensured.
This simplicity also facilitates extensions to other variants, such as non-Poissonian resetting~\cite{pal2016diffusion, nagar2016diffusion} and state-dependent reset probability~\cite{evans2011diffusionopt, roldan2017path}, etc.~\cite{evans2020stochastic}.
In particular, as illustrated by the U-shaped curves in Figs. 3–5, resetting can excessively constrain the target search of the SGD dynamics for sufficiently large values of $r$, which can degrade training efficiency. 
To address this, introducing occasional larger, more dynamic jumps (e.g., those inspired by L\`evy strategies and intermittent search strategies~\cite{benichou2011intermittent}) or adjusting the reset probability based on recent performance (e.g., decreasing the reset probability when no improvement is observed over time) can enhance the performance of our method while mitigating the risk of getting stuck.
Moreover, while we evaluated DNN models up to ResNet-34, we confirmed that the computational overhead, including I/O operations associated with resetting, is negligible compared to the overall training time. This makes our method scalable effectively to large-scale models, as further discussed in Sec.~D.8 of the SM.

The main limitation of this work is that the proposed method may not be as effective when the double descent phenomenon occurs or the convergence of validation loss is too late, such as the MAE case in Table~\ref{table:FinalTable}.
Moreover, it is challenging to identify the optimal reset probability from a limited number of experiments.
In fact, it has been observed that the coefficient of variation of the FPT is unity at the optimal reset probability in a random search problem~\cite{reuveni2016optimal, pal2017first}.
Future work investigating whether a similar relationship exists in DNN training may help to identify the optimal reset probability in practice.

We note that the resetting method shares a similar spirit with forgetting, which refers to the loss of previously acquired knowledge~\cite{wang2023comprehensive}.
Similar to resetting, forgetting was initially viewed as a catastrophic phenomenon that needed to be addressed~\cite{mccloskey1989catastrophic, parisi2019continual}; however, recent studies have highlighted its benefits, leading to its use in improving network performance~\cite{zhou2022fortuitous, nikishin2022primacy}.
From this perspective, resetting can be viewed as a form of forgetting memorized task-specific patterns, but unlike general forgetting that primarily erases early experiences, resetting targets the erasure of later experiences.
It will be interesting to further explore the connections between resetting and forgetting in future discussions.

We anticipate that our work can influence two different research directions.
First, it opens up the possibility of analyzing existing training methods from a statistical physics perspective, and second, it can pave the way to applying various new search strategies, beyond resetting, into neural network training.

\begin{ack}
This study was supported by the Basic Science Research Program through the National Research Foundation of Korea (Y.S. NRF Grant No. 2022R1A2B5B02001752, H.J. RS-2025-00514776). 
Y.B. was supported by an NRF grant funded by the Korean government (MSIT) (No. RS-2023-00278985).
\end{ack}

\vfill

\newpage
\bibliographystyle{ieeetr}
\bibliography{MLST-103133-R1}


\newpage
\appendix

\include{supple.tex}

\end{document}

%% file: supple.tex
\renewcommand\thefigure{S.\arabic{figure}}
\setcounter{figure}{0}

\renewcommand\thetable{S.\arabic{table}}
\setcounter{table}{0}

\renewcommand\theequation{S.\arabic{equation}}
\setcounter{equation}{0}

\section{SGD dynamics and Langevin dynamics}
\label{sec:appA}

\subsection{Derivation of Eq.~(2)}
We explain how the dynamics of SGD can be converted to the discretized Langevin equation.
We follow similar procedures as in Refs.~\cite{li2017stochastic, smith2018bayesian, ziyin2022strength}.
As written in the main text, the network parameters $\bm{\theta}$ are updated by
\begin{equation}
\begin{aligned}
    \Delta \bm{\theta}_t &= -\frac{\eta}{B}\sum_{(\bm{x}_i, \bm{y}_i)\in\mathcal{B}_t} \bm{\nabla}_{\bm{\theta}} \mathcal{L}_i(\bm{\theta}_t) \\
    & = -\frac{\eta}{N_{\rm tr }}\sum_{(\bm{x}_i, \bm{y}_i)\in\mathcal{D}_{\rm tr}} \bm{\nabla}_{\bm{\theta}} \mathcal{L}_i(\bm{\theta}_t) + \left( \frac{\eta}{N_{\rm tr }}\sum_{(\bm{x}_i, \bm{y}_i)\in\mathcal{D}_{\rm tr}} \bm{\nabla}_{\bm{\theta}} \mathcal{L}_i(\bm{\theta}_t) -\frac{\eta}{B}\sum_{(\bm{x}_i, \bm{y}_i)\in\mathcal{B}_{t}} \bm{\nabla}_{\bm{\theta}} \mathcal{L}_i(\bm{\theta}_t) \right) \\
    &= -\bm{\nabla}_{\bm{\theta}} \mathcal{R}_{\mathcal{D}_{\rm tr}}(\bm{\theta}_t) \eta + \bm{\xi}_t \sqrt{\eta},
\end{aligned}
\label{app_eq:SGD}
\end{equation}
where the random noise vector is $\bm{\xi}_t \equiv \sqrt{\eta} \left( \bm{\nabla}_{\bm{\theta}} \mathcal{R}_{\mathcal{D}_{\rm tr}}(\bm{\theta}_t) - \bm{\nabla}_{\bm{\theta}} \mathcal{R}_{\mathcal{B}_{t}}(\bm{\theta}_t) \right) \in \mathbb{R}^d$, $\bm{\nabla}_{\bm{\theta}} \mathcal{R}_{\mathcal{D}_{\rm tr}}(\bm{\theta}_t)$ is the gradient of risk on $\mathcal{D}_{\rm tr}$, and $\bm{\nabla}_{\bm{\theta}} \mathcal{R}_{\mathcal{B}_{t}}(\bm{\theta}_t)$ is the gradient of risk on a minibatch $\mathcal{B}_t$. 
Note that $\langle \bm{\nabla}_{\bm{\theta}}  \mathcal{L}_{i}(\bm{\theta}_t) \rangle_{\mathcal{D}_{\rm tr}} = \bm{\nabla}_{\bm{\theta}} \mathcal{R}_{\mathcal{D}_{\rm tr}}(\bm{\theta}_t)$ and $\langle \mathcal{R}_{\mathcal{B}_{t}}(\bm{\theta}_t) \rangle_{\mathcal{D}_{\rm tr}} = \mathcal{R}_{\mathcal{D}_{\rm tr}}(\bm{\theta}_t)$.
Using these facts, we can obtain that $\langle \bm{\xi}_t \rangle_{\mathcal{D}_{\rm tr}}=\bm{0}$ and
\begin{equation}
\begin{aligned}
    \langle \bm{\xi}_t \bm{\xi}_s^{\rm T} \rangle_{\mathcal{D}_{\rm tr}} &= \eta \left\langle \left( \bm{\nabla}_{\bm{\theta}} \mathcal{R}_{\mathcal{D}_{\rm tr}}(\bm{\theta}_t) - \bm{\nabla}_{\bm{\theta}} \mathcal{R}_{\mathcal{B}_{t}}(\bm{\theta}_t) \right) \left( \bm{\nabla}_{\bm{\theta}} \mathcal{R}_{\mathcal{D}_{\rm tr}}(\bm{\theta}_s) - \bm{\nabla}_{\bm{\theta}} \mathcal{R}_{\mathcal{B}_{s}}(\bm{\theta}_s) \right)^{\rm T} \right\rangle_{\mathcal{D}_{\rm tr}} \\
    &= \eta \left( \left\langle \bm{\nabla}_{\bm{\theta}} \mathcal{R}_{\mathcal{B}_{t}}(\bm{\theta}_t) \bm{\nabla}_{\bm{\theta}} \mathcal{R}_{\mathcal{B}_{s}}(\bm{\theta}_s)^{\rm T}\right\rangle_{\mathcal{D}_{\rm tr}} - \bm{\nabla}_{\bm{\theta}} \mathcal{R}_{\mathcal{D}_{\rm tr}}(\bm{\theta}_t) \bm{\nabla}_{\bm{\theta}} \mathcal{R}_{\mathcal{D}_{\rm tr}}(\bm{\theta}_s)^{\rm T}  \right).
\end{aligned}
\end{equation}
For the $s \neq t$ case, $\left\langle \bm{\nabla}_{\bm{\theta}} \mathcal{R}_{\mathcal{B}_{t}}(\bm{\theta}_t) \bm{\nabla}_{\bm{\theta}} \mathcal{R}_{\mathcal{B}_{s}}(\bm{\theta}_s)^{\rm T}\right\rangle_{\mathcal{D}_{\rm tr}} = \bm{\nabla}_{\bm{\theta}} \mathcal{R}_{\mathcal{D}_{\rm tr}}(\bm{\theta}_t) \bm{\nabla}_{\bm{\theta}} \mathcal{R}_{\mathcal{D}_{\rm tr}}(\bm{\theta}_s)^{\rm T}$ because each minibatch is independently sampled from $\mathcal{D}_{\rm tr}$.
Applying $\langle \bm{\nabla}_{\bm{\theta}} \mathcal{L}_{i}(\bm{\theta}_t) \bm{\nabla}_{\bm{\theta}} \mathcal{L}_{j}(\bm{\theta}_t)\rangle_{\mathcal{D}_{\rm tr}} = \| \bm{\nabla}_{\bm{\theta}} \mathcal{R}_{\mathcal{D}_{\rm tr}}(\bm{\theta}_t) \|^2 + \mathsf{\Sigma}(\bm{\theta}_t)\delta_{ij}$ with the covariance matrix $\mathsf{\Sigma}(\bm{\theta}_t)$, where $\| \cdot \|$ denotes the Euclidean norm of a vector, we have
\begin{equation}
\begin{aligned}
    \left\langle \| \bm{\nabla}_{\bm{\theta}} \mathcal{R}_{\mathcal{B}_{t}}(\bm{\theta}_t) \|^2\right\rangle_{\mathcal{D}_{\rm tr}} = \| \bm{\nabla}_{\bm{\theta}} \mathcal{R}_{\mathcal{D}_{\rm tr}}(\bm{\theta}_t) \|^2 + \frac{1}{B} \mathsf{\Sigma}(\bm{\theta}_t).
\end{aligned}
\end{equation}
Therefore, we obtain $\langle \bm{\xi}_t \bm{\xi}_s^{\rm T} \rangle_{\mathcal{D}_{\rm tr}}= 2\mathsf{D}(\bm{\theta}_t) \delta_{ts}$ with $\mathsf{D}(\bm{\theta}_t)=\eta \mathsf{\Sigma}(\bm{\theta}_t)/(2B)$. 
We assumed $N_{\rm tr} \gg B$ in the above derivation, but the decreasing trend of $\mathsf{D}(\bm{\theta}_t)$ with $B$ is still valid for $N_{\rm tr} \geq B$~\cite{ziyin2022strength}.
The noise covariance matrix $\mathsf{D}$ is typically state-dependent and highly anisotropic, which contributes to the heavy-tailed and nonequilibrium stationary distribution of the SGD dynamics~\cite{chaudhari2018stochastic, gurbuzbalaban2021heavytail}. 
In addition, it has been shown that $\mathsf{D}$ is positively correlated with the Hessian matrix of the loss~\cite{jastrzebski2017three, li2020hessian}, indicating that the noise strength is greater in directions where the loss landscape is sharper. This relationship suggests that SGD prefers flat minima over sharp minima~\cite{xie2021diffusion, yang2023stochastic}.

Let us consider the overdamped Langevin equation, a first-order stochastic differential equation describing the evolution of a particle where friction dominates over inertia.
Applying the Euler method, the overdamped Langevin equation can be approximated with time interval $\Delta t$ by~\cite{gardiner2004handbook}
\begin{equation}
\begin{aligned}
    \Delta \bm{x}_t = -\bm{\nabla}_{\bm{x}} V(\bm{x}_t) \Delta t + \sqrt{2 \mathsf{B}(\bm{x}_t) } \Delta \bm{W}_t.
\end{aligned}
\label{app_eq:OLE}
\end{equation}
Here, $\bm{x}_t$ is the position of the particle at time step $t$, $V(\bm{x}_t)$ is the underlying potential, $\mathsf{B}(\bm{x}_t)$ is the strength of the fluctuations, called the diffusion matrix, and $\Delta \bm{W}_t$ is the random noise vector that satisfies $\langle \Delta \bm{W}_t \rangle = \bm{0}$ and $\langle \Delta \bm{W}_t \Delta \bm{W}_s^{\rm T}\rangle = \Delta t \delta_{ts}$, where $\langle \cdot \rangle$ denotes the ensemble average.
When we simulate Eq.~\eqref{app_eq:OLE}, we randomly sample the random real number $\bm{\zeta}_t$ from a certain probability distribution with zero-mean and unit-variance at each iteration $t$ and represent the noise vector as $\Delta \bm{W}_t \equiv \bm{\zeta}_t \sqrt{\Delta t}$.
Note that $\Delta \bm{W}_t$ is commonly assumed as an increment of the Wiener process based on the central limit theorem, so that $\bm{\zeta}_t$ is generally sampled from the standard normal distribution. This assumption is often violated in various situations~\cite{metzler2014anomaloous}, and $\bm{\zeta}_t$ can be sampled from other distributions depending on the system.
Comparing Eq.~\eqref{app_eq:SGD} with Eq.~\eqref{app_eq:OLE}, we can easily see that the dynamics of SGD follows the overdamped Langevin equation, with $\mathcal{R}_{\mathcal{D}_{\rm tr}}$ and $\mathsf{D}$ serving as the potential and the diffusion matrix, respectively.
Based on this correspondence, we analyze the SGD dynamics in the language of the Langevin dynamics and apply the stochastic resetting strategy in the main text.

\subsection{SGD dynamics for noisy small dataset}

In Sec. 3, we derived that latent gradient bias arises caused by label noise hinders the SGD dynamics from reaching the optimal parameters.
While the main text primarily focuses on the impact of label noise, other factors can similarly degrade the training of DNNs.
One such factor is the limited size of the training dataset. To investigate how dataset size affects DNN training, we recall the equation governing SGD dynamics with label noise [Eq.~(3) in the main text]:
\begin{equation}
\begin{aligned}
    \Delta \bm{\theta}_t = -\left[ \bm{\nabla}_{\bm{\theta}} \hat{\mathcal{R}}_{\tilde{\mathcal{D}}_{\rm tr}^c}(\bm{\theta}_t) + \bm{\nabla}_{\bm{\theta}} \hat{\mathcal{R}}_{\tilde{\mathcal{D}}_{\rm tr}^w}(\bm{\theta}_t) \right] \eta + \bm{\xi}_t \sqrt{\eta},
\end{aligned}
\label{app_eq:SGD_LE2}
\end{equation}
with the label noise rate $0 \leq \tau \leq 1$ is defined, the gradients from the correct part $\bm{\nabla}_{\bm{\theta}} \hat{\mathcal{R}}_{\tilde{\mathcal{D}}_{\rm tr}^c}(\bm{\theta}_t) \equiv (1-\tau)\bm{\nabla}_{\bm{\theta}} {\mathcal{R}}_{\tilde{\mathcal{D}}_{\rm tr}^c}(\bm{\theta}_t)$, and the gradients from the wrong part $\bm{\nabla}_{\bm{\theta}} \hat{\mathcal{R}}_{\tilde{\mathcal{D}}_{\rm tr}^w}(\bm{\theta}_t) \equiv \tau\bm{\nabla}_{\bm{\theta}} {\mathcal{R}}_{\tilde{\mathcal{D}}_{\rm tr}^w}(\bm{\theta}_t)$. 
For a sufficiently large training dataset $\mathcal{D}_{\rm tr}$, $\bm{\nabla}_{\bm{\theta}}{\mathcal{R}}_{\tilde{\mathcal{D}}_{\rm tr}^c}(\bm{\theta}_t)$ closely approximates the gradients of the true risk, denoted by $\bm{\nabla}_{\bm{\theta}}{\mathcal{R}}_{{\mathcal{D}}}(\bm{\theta}_t)$.
However, when $\mathcal{D}_{\rm tr}$ is small, a discrepancy between them emerges due to the small dataset size, which can be expressed by
\begin{equation}
\begin{aligned}
    \bm{\nabla}_{\bm{\theta}} \hat{\mathcal{R}}_{\tilde{\mathcal{D}}_{\rm sm}^c}(\bm{\theta}_t) =    
    \bm{\nabla}_{\bm{\theta}} \hat{\mathcal{R}}_{\tilde{\mathcal{D}}_{\rm tr}^c}(\bm{\theta}_t) -
    \bm{\nabla}_{\bm{\theta}} \hat{\mathcal{R}}_{\mathcal{D}}(\bm{\theta}_t),
\end{aligned}
\label{app_eq:bias_small}
\end{equation}
with $\bm{\nabla}_{\bm{\theta}} \hat{\mathcal{R}}_{\mathcal{D}}(\bm{\theta}_t) \equiv (1-\tau) \bm{\nabla}_{\bm{\theta}}{\mathcal{R}}_{\mathcal{D}}(\bm{\theta}_t)$.
Substituting Eq.~\eqref{app_eq:bias_small} into Eq.~\eqref{app_eq:SGD_LE2} yields:
\begin{equation}
\begin{aligned}
    \Delta \bm{\theta}_t = -\bm{\nabla}_{\bm{\theta}} \hat{\mathcal{R}}_{\mathcal{D}}(\bm{\theta}_t) \eta - \bm{\nabla}_{\bm{\theta}} \hat{\mathcal{R}}_{\tilde{\mathcal{D}}_{\rm sm}^c}(\bm{\theta}_t) \eta - \bm{\nabla}_{\bm{\theta}} \hat{\mathcal{R}}_{\tilde{\mathcal{D}}_{\rm tr}^w}(\bm{\theta}_t) \eta + \bm{\xi}_t \sqrt{\eta}.
\end{aligned}
\label{app_eq:SGD_LE3}
\end{equation}
Here, the first term in the right-hand-side of Eq.~\eqref{app_eq:SGD_LE3} represents the drift toward the true optimum, the second term represents the bias caused by the small dataset size, and the third term represents the bias introduced by label noise.
Thus, we can identify that latent gradient bias arises from either a small dataset size or label noise, both of which hinder the generalization ability of DNN.
Notably, for a large $\mathcal{D}_{\rm tr}$, the second term $\bm{\nabla}_{\bm{\theta}} \hat{\mathcal{R}}_{\tilde{\mathcal{D}}_{\rm sm}^c}(\bm{\theta}_t)$ diminishes, leading to $\bm{\nabla}_{\bm{\theta}} \hat{\mathcal{R}}_{\mathcal{D}}(\bm{\theta}_t) \simeq \bm{\nabla}_{\bm{\theta}} \hat{\mathcal{R}}_{\tilde{\mathcal{D}}_{\rm tr}^c}(\bm{\theta}_t)$.
On the other hand, for $\tau = 0$ case, the third term $\bm{\nabla}_{\bm{\theta}} \hat{\mathcal{R}}_{\tilde{\mathcal{D}}_{\rm tr}^w}(\bm{\theta}_t)$ disappears, and the first and second terms dominate. Conversely, for $\tau = 1$, the third term persists, and the first and second terms vanish.
These results suggest that the label noise rate influences the relative magnitudes of the drift terms, whereas the dataset size affects the direction of the drift by determining how effectively the correct labels within the dataset guide the model toward the true optimum.
Our analysis mainly centers on the effects of label noise, but exploring the individual and combined impacts of $\bm{\nabla}_{\bm{\theta}} \hat{\mathcal{R}}_{\tilde{\mathcal{D}}_{\rm sm}^c}(\bm{\theta}_t)$ on DNN training would be an intriguing direction for future work.
Furthermore, the substantial improvements achieved by restarting with a small noisy dataset, as demonstrated in Secs. 4.1–3, highlight the efficacy of our method in this challenging regime.

\section{Stochastic resetting in statistical physics}
\label{sec:appB}

We briefly introduce what stochastic resetting is and the conditions under which this strategy can help random searches (see Ref.~\cite{evans2020stochastic} for a more detailed review).
In fact, the resetting method has already been exploited in some stochastic algorithms~\cite{villen1991restart, luby1993optimal, tong2008random}, but much attention has been attracted by the theoretical success of stochastic resetting~\cite{evans2011diffusion}.
Several approaches have been made to deal with search processes involving resets, such as calculating the survival probability~\cite{evans2011diffusion, ray2019peclet}.
Here, we present an easy but general one, and clarify that we follow the same procedure as in Refs.~\cite{reuveni2016optimal, pal2017first, pal2020search}.

Let us consider a generic searcher that starts from a resetting point at time zero in a $d$-dimensional space, with the assumption that the searcher resets at a rate of $\gamma$ if the target is not found.
In other words, the search process is completed when the searcher finds the target before resetting; otherwise, the searcher returns to the resetting point and repeats this procedure until the target is found.
This procedure can be understood in terms of two random variables, $T$ and $R$, which represent the time to find a target and the time to reset, respectively. 
If we draw $T$ and $R$ from their respective distributions, we check whether $T > R$. If $T > R$, the searcher resets before finding the target, and the search process begins anew from the resetting point. Conversely, if $T < R$, the searcher finds the target before resetting, and the search process is completed.
Applying this scheme, the time to find a target, known as the first passage time (FPT) and denoted by $T(\gamma)$, can be expressed by the following renewal equation:
\begin{equation}
\begin{aligned}
    T(\gamma) = 
    \begin{cases}
        T, &\text{if $T<R$}, \\
        R + T(\gamma)',
        &\text{if $R \leq T$,}
    \end{cases}
\end{aligned}
\end{equation}
or equivalently,
\begin{equation}
\begin{aligned}
    T(\gamma) = \min (T, R) + I(R \leq T) T(\gamma)',
\end{aligned}
\label{app_eq:renewal2}
\end{equation}
where $T(\gamma)'$ denotes an independent and identically distributed copy of $T(\gamma)$, and $I(R\leq T)$ denotes an indicator function which equals one if $R \leq T$ and zero otherwise.
Note that $T(\gamma) = T$ if there is no reset ($\gamma = 0$).
Taking expectations in Eq.~\eqref{app_eq:renewal2}, we obtain the mean first passage time (MFPT):
\begin{equation}
\begin{aligned}
    \langle T(\gamma) \rangle = \frac{\langle \min (T, R) \rangle}{{\rm Pr}(T < R)},
\end{aligned}
\label{app_eq:MFPT1}
\end{equation}
with the relations $\langle I(R\leq T) \rangle = {\rm Pr}(R \leq T)$ and $\langle T(\gamma) \rangle = \langle T(\gamma)' \rangle$.
It is noteworthy that we do not assume the dynamics of the search process and the function of the reset rate.
Thus, Eq.~\eqref{app_eq:MFPT1} can be applied to a general search process regardless of the distributions of $T$ and $R$.

As a simple reset method, we assume a constant reset rate, i.e., the reset probability within a time interval $dt$ is $\gamma dt$~\cite{evans2011diffusion}.
Then, the distribution of $R$ is exponential with $\gamma e^{-\gamma t}$ at time $t$ and $\langle \min (T, R) \rangle$ can be calculated by
\begin{equation}
\begin{aligned}
    \langle \min (T, R) \rangle &= \int_{0}^{\infty} dt \left[ 1- {\rm Pr}(\min (T, R) \leq t) \right] \\
    &= \int_{0}^{\infty} dt \;{\rm Pr}(R>t) {\rm Pr}(T>t)  \\
    &= \int_{0}^{\infty} dt \; e^{-rt} \int_{t}^{\infty} dt' f_T(t') = \frac{1}{r} - \frac{1}{r} \int_{0}^{\infty} dt \; e^{-rt}f_T(t).
\end{aligned}
\end{equation}
In addition, ${\rm Pr}(T < R) = \int_{0}^{\infty}dt \; f_T(t) {\rm Pr}(R > t)=\int_{0}^{\infty} dt \; e^{-rt}f_T(t)$.
Substituting these equations into Eq.~\eqref{app_eq:MFPT1}, we obtain
\begin{equation}
\begin{aligned}
    \langle T(\gamma) \rangle = \frac{1-\tilde{T}(\gamma)}{\gamma \tilde{T}(\gamma)}
\end{aligned}
\label{app_eq:MFPT2}
\end{equation}
where $\tilde{T}(\gamma) \equiv \int_{0}^{\infty} dt e^{-\gamma t} f_T(t)$ denotes the Laplace transform of $T$ evaluated at $\gamma$.
Note that $f_T(t)$ is determined by the underlying dynamics of the searcher, which may include factors such as stochasticity or external drift.
Therefore, once the dynamics of a searcher are determined and $f_T(t)$ is known, we can calculate the MFPT at $\gamma$ and identify whether resetting is beneficial.

To obtain the MFPT with resetting, let us specify the dynamics of a searcher. 
Consider a searcher diffusing in one dimension with a diffusion constant $D$ and a constant drift $v$ and assume that the searcher starts at the origin (reset point) and that the target we want to find is located at $L$ ($>0$).
Here, $D$ represents the stochasticity and $v$ represents the drift toward a target.
The position $x(t)$ of the searcher at time $t$ evolves during a small time interval $\Delta t$ through the Langevin equation given by
\begin{equation}
\begin{aligned}
    x (t+\Delta t) = 
    \begin{cases}
        x_r, &\text{with probability $\gamma \Delta t$}, \\
        x(t) + v\Delta t + \xi(t) \sqrt{\Delta t},
        &\text{otherwise,}
    \end{cases}
\end{aligned}
\label{app_eq:LE}
\end{equation}
where $x_r$ denotes the reset point set as the origin and $\xi(t)$ denotes a stochastic force, typically modeled Gaussian white noise satisfying $\langle \xi(t) \rangle = 0$ and $\langle \xi(t) \xi(s) \rangle = 2D \delta(t-s)$. 
In this search process, the probability distribution to find the searcher at position $x$ at time $t$ is known to be given by~\cite{gardiner2004handbook}
\begin{equation}
\begin{aligned}
    G_0(x, t) = \frac{1}{\sqrt{4 \pi Dt}}\left[ e^{-\frac{(x-vt)^2}{4Dt}} - e^{\frac{Lv}{D}}e^{-\frac{(x-2L-vt)^2}{4Dt}} \right].
\end{aligned}
\label{app_eq:propagator}
\end{equation}
The FPT distribution $f_T(t)$ is then derived from the probability that the searcher has not yet reached the target by time $t$: $f_T(t) = \frac{d}{dt} {\rm Pr}(T \geq t) = \frac{d}{dt} \int_{-\infty}^{L} dx \, G_0 (x, t)$.
Using this equation and the definition of $\tilde{T}(\gamma)$, we find $\tilde{T}(\gamma)=1-\gamma \int^{L}_{-\infty} dx \, \tilde{G}_0(x, \gamma)$ where $\tilde{G}_0 (x, \gamma) \equiv \int_{0}^{\infty} dt e^{-\gamma t} G_0(x, t)$ is the Laplace transform of $G_0(x, t)$ evaluated at $\gamma$.
Thus, upon calculation with these equations and Eq.~\eqref{app_eq:propagator}, we obtain
\begin{equation}
\begin{aligned}
    \tilde{T}(\gamma) = e^{\frac{L}{2D} \left(v - \sqrt{v^2+4D\gamma} \right)}.
\end{aligned}
\label{app_eq:LaplaceT}
\end{equation}
Substituting Eq.~\eqref{app_eq:LaplaceT} into Eq.~\eqref{app_eq:MFPT2}, we finally have
\begin{equation}
\begin{aligned}
    \langle T(\gamma) \rangle = \frac{1}{\gamma} \left[ e^{ \frac{L}{2D} \left(\sqrt{v^2 + 4D\gamma} - v \right)}  - 1\right].
\end{aligned}
\label{app_eq:MFPT_drift}
\end{equation}
This final expression gives the MFPT for a searcher with drift and diffusion, including the effect of resetting at a rate $\gamma$.

In our paper, we exploit the correspondence between SGD dynamics and Langevin dynamics, as described in Sec.~\ref{sec:appA}, and theoretically identify where stochasticity is strengthened and drift toward a target is weakened in DNN training.
Although we demonstrated a simple one-dimensional case in this section, numerous studies have explored more complex scenarios, including various confining potentials~\cite{pal2015diffusion, pal2020search}, and related contexts.~\cite{evans2020stochastic}.
Importantly, it has been proven that a diffusive particle with resetting converges to a nonequilibrium steady state even in arbitrary spatial dimensions, and it has also been shown to be beneficial for target search~\cite{evans2011diffusionopt, evans2014diffusion}.
We believe these findings support the hypothesis that resetting may provide advantages in the search process involved in SGD dynamics for DNN.
However, we also acknowledge that the advantages of resetting in terms of MFPT are not directly connected to performance improvements in DNN training.
We conjecture and empirically validate that similar beneficial mechanisms successfully operate in DNN training. 
Nonetheless, it would be necessary and intriguing to find an alternative metric to represent the DNN performance and to theoretically investigate the effect of resetting.

\section{Description of the experiments}

This section provides details on the experiments not included in the main text. For all of the experiments, we perform 5 independent runs to achieve the average and the standard error values. All runs were made independently on a single NVIDIA TITAN V GPU. 
All results are obtained from the model at the optimal iteration based on minimum validation loss as default, except ELR and SOP+ methods.
Additionally, the resulting test accuracy is evaluated on the clean validation set.
The objective throughout our experiments is to compare the performance with and without resetting, not to achieve state-of-the-art performances; therefore, we did not heavily tune the hyperparameters for each of the settings. Here we provide the choice of hyperparameters for our experiments.

\subsection{Network architecture}

We employed a vanilla CNN (VCNN) as mentioned in Secs.~4.1, 4.2, and 4.3 to expedite a straightforward testing of our claims.
It consists of simple layers, as outlined in Table~\ref{table:VCNN}.
In the table, the inclusion of batch normalization before the activation function and after a layer is indicated by the term "Use BatchNorm". 
The output dimension of a convolutional layer is represented as $(C, W, H)$, where $C$ denotes the number of channels, and $W$ and $H$ represent the width and height, respectively. Additionally, $c$ represents the number of classes in the dataset.

\begin{table}[H]
\caption{Network architecture of the VCNN: Layer name, output dimension of the layer, parameters of the convolutional layer $(K, P, S)$, and activation function. Here, $K$, $P$, and $S$ represent the size of the filter, padding, and stride, respectively.}
\centering
\begin{tabular}{c|c|c|c|c}
\hline
\rule{0pt}{2.0ex}
Layer name & Output dim. & $(K, P, S)$ & Use BatchNorm & Activation function\\
\hline
\rule{0pt}{2.0ex}
Input image & $\left( 3, 32, 32\right)$ &  None & X & None \\
Conv2d & $\left( 32, 32, 32 \right)$ & $(3, 1, 1)$ & O & ReLU \\
Conv2d & $\left( 64, 32, 32 \right)$ & $(3, 1, 1)$ & O & ReLU\\
MaxPool2d & $\left( 64, 16, 16 \right)$ & $(2, 0, 2)$ & O & None\\
Conv2d & $\left( 128, 16, 16 \right)$ & $(3, 1, 1)$ & O & ReLU \\
Conv2d & $\left( 128, 16, 16 \right)$ & $(3, 1, 1)$ & O & ReLU\\
MaxPool2d & $\left( 128, 8, 8 \right)$ & $(2, 0, 2)$ & X & None\\
Conv2d & $\left( 256, 8, 8 \right)$ & $(3, 1, 1)$ & O & ReLU \\
Conv2d & $\left( 256, 8, 8 \right)$ & $(3, 1, 1)$ & O & ReLU\\
MaxPool2d & $\left( 256, 4, 4 \right)$ & $(2, 0, 2)$ & X & None\\
Dropout ($p=0.2$) & $256 \times 4 \times 4$ & None & X & None\\
Linear & $1024$ & None & O & ReLU\\
Linear & $512$ & None & O & ReLU\\
Linear & $c$ & None & X & Softmax\\
\hline
\end{tabular}
\label{table:VCNN}
\end{table}

\vfill
\newpage

\subsection{Experimental setting for results on benchmark datasets in Secs.~4.4 and~4.5}

We set the hyperparameters required for each baseline method in Secs.~4.4 and~4.5 following Table~\ref{table:hyper-settings}, as specified in their original papers.

\begin{table}[H]
\caption{Hyperparameters for baseline methods used in Secs.~4.4 and~4.5.}
\centering
\begin{tabular}{ccc}
\hline
\rule{0pt}{2.0ex}
\multirow{2}{*}{Method}          & \multicolumn{2}{c}{Dataset}                                               \\
                                 & CIFAR-10/10N                        & CIFAR-100/100N                      \\ \hline
\rule{0pt}{2.0ex}
GCE~\cite{zhang2018generalized}  & $(q, k)=(0.7, 0)$                   & $(q, k)=(0.7, 0)$                   \\
SL~\cite{wang2019symmetric}      & $(\alpha, \beta)=(0.1, 1.0)$        & $(\alpha, \beta)=(6.0, 0.1)$        \\
ELR~\cite{liu2020early-learning} & $\lambda = 3$                       & $\lambda = 7$                       \\
ELR*                             & $\lambda = 0.5$                     & $\lambda = 1$                       \\
SOP+\cite{liu2022robust}         & $(\alpha_u , \alpha_v ) = (10, 10)$ & $(\alpha_u , \alpha_v ) = (1, 10)$  \\
SOP+*                            & $(\alpha_u , \alpha_v ) = (1, 1)$   & $(\alpha_u , \alpha_v ) = (0.1, 1)$ \\ \hline
\end{tabular}
\label{table:hyper-settings}
\end{table}

We confirm that all results for methods without our proposed approach are consistent with the performance reported in their original papers.
We set the batch size to 256, momentum to $0.9$, and weight decay to $5 \times 10^{-4}$ for the cross-entropy loss, MAE, GCE, and SL methods (for SL on CIFAR-100, the batch size is set to 128). For the ELR and SOP methods, we set the batch size to 128, momentum to $0.9$, and weight decay to $10^{-3}$, following Ref.~\cite{liu2020early-learning, liu2022robust}.
We employ a cosine annealing scheduler~\cite{loshchilov2017sgdr}, setting the maximum number of iterations to the total iteration $5 \times 10^4$ with an initial learning rate of $0.1$ for the cross-entropy loss and $0.01$ for MAE~\cite{ghosh2017robust}, GCE, and SL methods.
For the ELR and SOP+ methods, we set the initial learning rate as $0.02$, and reduce it by $1/100$ and $1/10$, respectively, after 40 (80) and 80 (120) with a total epoch 150 on CIFAR-10 (CIFAR-100) as indicated in Ref.~\cite{liu2020early-learning, liu2022robust}. Additional regularizer parameters for SOP+ and SOP+* are $(\lambda_C, \lambda_B) = (0.9, 0.1)$.
Moreover, for the ELR and SOP+ methods, we set the reset checkpoint and select the best model based on validation accuracy instead of validation loss due to the discrepancy in the resulting test accuracies between these accuracy-based and loss-based approaches in the baseline. The threshold $\mathcal{T}$ is set to $30$ epochs on ELR and SOP+. Especially in SOP+, our method starts after $80$ epochs on CIFAR-100 (Table~1).

When testing on ViT~\cite{dosovitskiy2021an}, we use the ViT-Tiny (ViT-T) model from the Hugging Face timm library~\cite{rw2019timm}. The model uses a patch size of 16, an embedding dimension of 192, a depth of 12, and 3 attention heads, with an input image size of 224.
For training ViT-T, we use cross-entropy loss and SGD with a momentum of 0.9 and no weight decay. The batch size is set to 256, and the initial learning rate is 0.01. We employ a cosine annealing scheduler with the maximum number of iterations to $5 \times 10^4$.
For ANIMAL-10N, we followed the default setting of cross-entropy loss, modifying only the batch size to 128.

\vfill

\section{Additional results}
\subsection{Cosine similarities between drifts in Sec.~3}
\begin{figure}[!ht]
    \includegraphics{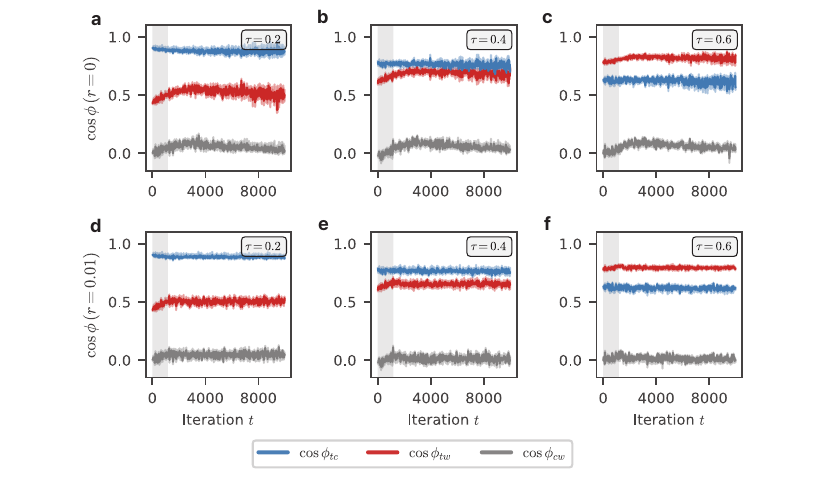}
    \vskip -0.1in
    \caption{Cosine similarities between $-\bm{\nabla}_{\bm{\theta}} \mathcal{R}_{\tilde{\mathcal{D}}_{\rm tr}}(\bm{\theta})$ (total drift), $-\bm{\nabla}_{\bm{\theta}} \hat{\mathcal{R}}_{\tilde{\mathcal{D}}^c_{\rm tr}}(\bm{\theta})$ (drift from the correct label, i.e., correct drift), and $-\bm{\nabla}_{\bm{\theta}} \hat{\mathcal{R}}_{\tilde{\mathcal{D}}^w_{\rm tr}}(\bm{\theta})$ (drift from the wrong label, i.e., wrong drift). The first row (a, b, c) shows the cosine similarities when there is no reset, $r=0$, and the second row (d, e, f) shows the cosine similarities when resetting is applied, $r=0.01$, with (a, d) a noise rate of $\tau=0.2$, (b, e) $\tau=0.4$, and (c, f) $\tau=0.6$. Here, $\cos \phi_{tc}$, $\cos \phi_{tw}$, and $\cos \phi_{cw}$ denote the cosine similarity between total and correct drifts, total and wrong drifts, and correct and wrong drifts, respectively. The gray region represents the iterations before the resetting commences. We set $B=8$ in Setting 1 and performed 5 independent runs, where the shaded areas denote the standard error.
    }\label{supp1}
\end{figure}

We provide the cosine similarities between drift components, i.e., $-\bm{\nabla}_{\bm{\theta}} \mathcal{R}_{\tilde{\mathcal{D}}_{\rm tr}}(\bm{\theta})$ (total drift), $-\bm{\nabla}_{\bm{\theta}} \hat{\mathcal{R}}_{\tilde{\mathcal{D}}^c_{\rm tr}}(\bm{\theta})$ (drift from the correct part), and $-\bm{\nabla}_{\bm{\theta}} \hat{\mathcal{R}}_{\tilde{\mathcal{D}}^w_{\rm tr}}(\bm{\theta})$ (drift from the wrong part, i.e., latent gradient bias by label noise) by varying the noise rate in Fig.~\ref{supp1}.
The shaded gray area represents the iteration region before the stochastic resetting takes place, and the white area represents the iteration region after the resetting commences. 
Note that the stochastic resetting does not directly change the gradient itself, as can be seen by comparing the top row (no reset) in Fig.~\ref{supp1}(a, b, c) and the bottom row (with reset) in Fig.~\ref{supp1}(d, e, f). Instead, stochastic resetting suppresses trajectories of DNNs that drift away from the optimum, thereby increasing the chances of finding it despite the presence of latent gradient bias.
Here, we can see that the cosine similarity between $-\bm{\nabla}_{\bm{\theta}} \hat{\mathcal{R}}_{\tilde{\mathcal{D}}_{\rm tr}^c}(\bm{\theta}_t)$ and $-\bm{\nabla}_{\bm{\theta}} \hat{\mathcal{R}}_{\tilde{\mathcal{D}}_{\rm tr}^w}(\bm{\theta}_t)$, denoted by $\cos \phi_{cw}$, is almost zero, indicating that $-\bm{\nabla}_{\bm{\theta}} \hat{\mathcal{R}}_{\tilde{\mathcal{D}}_{\rm tr}^c}(\bm{\theta}_t)$ and $-\bm{\nabla}_{\bm{\theta}} \hat{\mathcal{R}}_{\tilde{\mathcal{D}}_{\rm tr}^w}(\bm{\theta}_t)$ are most likely to be orthogonal to each other (see the further discussion in Sec.~\ref{sec:appD_orthogonal}).
In addition, the cosine similarity between $-\bm{\nabla}_{\bm{\theta}} {\mathcal{R}}_{\tilde{\mathcal{D}}_{\rm tr}}(\bm{\theta}_t)$ and $-\bm{\nabla}_{\bm{\theta}} \hat{\mathcal{R}}_{\tilde{\mathcal{D}}_{\rm tr}^c}(\bm{\theta}_t)$, denoted by $\cos \phi_{tc}$, decreases with increasing $\tau$, implying that $-\bm{\nabla}_{\bm{\theta}} {\mathcal{R}}_{\tilde{\mathcal{D}}_{\rm tr}}(\bm{\theta}_t)$ becomes less correlated with $-\bm{\nabla}_{\bm{\theta}} \hat{\mathcal{R}}_{\tilde{\mathcal{D}}_{\rm tr}^c}(\bm{\theta}_t)$.
These results support our claims in Sec.~3 and the schematic in Fig.~2(a).

\newpage

\subsection{Orthogonality between the drift components}
\label{sec:appD_orthogonal}

\begin{figure}[!ht]
    \vskip -0.1in
    \includegraphics{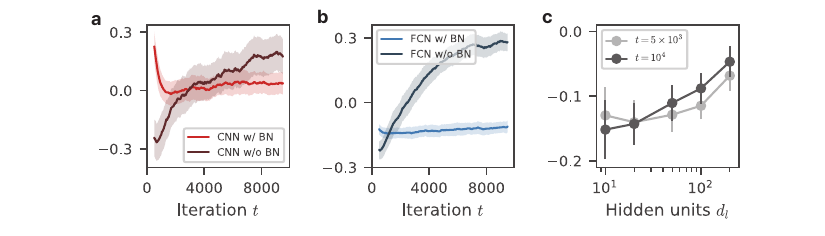}
    \vskip -0.1in
    \caption{Cosine similarities between $-\bm{\nabla}_{\bm{\theta}} \hat{\mathcal{R}}_{\tilde{\mathcal{D}}^c_{\rm tr}}(\bm{\theta})$ and $-\bm{\nabla}_{\bm{\theta}} \hat{\mathcal{R}}_{\tilde{\mathcal{D}}^w_{\rm tr}}(\bm{\theta})$, denoted by $\cos \phi_{cw}$, during training with (w/) and without (w/o) batch normalization (BN) for (a) CNN and (b) FCN structures.
    (c) $\cos \phi_{cw}$ by varying the number of hidden units $d_l$ in the FCN structure with batch normalization.
    We set $B=8$ in Setting 1 and performed 5 independent runs, where the shaded areas denote the standard error.
    Additionally, we perform a moving average with a time window of $50$ log-iterations to smooth the recorded values of $\cos \phi_{cw}$, which are measured every 20 iterations.
    }\label{supp2}
\end{figure}

We further explore the orthogonality of $-\bm{\nabla}_{\bm{\theta}} \hat{\mathcal{R}}_{\tilde{\mathcal{D}}_{\rm tr}^c}(\bm{\theta}_t)$ and $-\bm{\nabla}_{\bm{\theta}} \hat{\mathcal{R}}_{\tilde{\mathcal{D}}_{\rm tr}^w}(\bm{\theta}_t)$. 
In contrast to our results, Ref.~\cite{feng2021phases} demonstrated that $\cos \phi_{cw}$ varies significantly during training and can be used as an order parameter to divide the learning phases.
To investigate the factors contributing to this discrepancy, we perform additional experiments with DNNs of comparable sizes and settings to those used in Ref.~\cite{feng2021phases}, as detailed in Table~\ref{table:CNN_FCN}.
Note that for the fully connected network (FCN) structures, the case with $d_l = 50$ number of hidden units matches the configuration used in Ref.~\cite{feng2021phases}.
A key difference in our setup is the use of batch normalization~\cite{ioffe2015batch}, which we hypothesize may play a significant role in the observed effects.
Interestingly, as shown in Fig.~\ref{supp2}, we observe that batch normalization enforces orthogonality between two drift components during training in both CNN and FCN structures.
Moreover, this orthogonality in the presence of batch normalization becomes more pronounced as the number of hidden units $d_l$ increases, implying that the dimensionality of network parameters also affects the orthogonality [Fig.~\ref{supp2}(b,c)].
These observations are in a similar line as findings from Ref.~\cite{daneshmand2021batch}, which reported that batch normalization in DNNs induces orthogonalization of hidden representations of samples across layers.
As batch normalization is one of the most widely adopted techniques for improving training in DNNs, our findings suggest that the orthogonality between the drift components is likely to generalize across diverse DNN structures and experimental settings.
Exploring the mechanisms by which batch normalization enforces this orthogonality would be an intriguing direction for future research, complementing theoretical studies on batch normalization, such as those in Refs.~\cite{yang2018meanfield, daneshmand2021batch, Joudaki2023on_bridging}.

\begin{table}[h]
     \caption{Network architectures of (top) CNN and (bottom) FCN used in Sec.~\ref{sec:appD_orthogonal}. Here, $d_l$ represents the number of hidden units of the FCN structure.
     Batch normalization, when applied, is placed before the activation function, except for the output layer.}
     \centering
    \begin{subtable}
        \centering
        \begin{tabular}{c|c|c|c}
        \multicolumn{4}{c}{CNN} \\
        \hline
        \rule{0pt}{2.0ex}
        Layer name & Output dim. & $(K, P, S)$ & Activation function \\
        \hline
        \rule{0pt}{2.0ex}
        Input image & $\left( 3, 32, 32\right)$ &  None & None \\
        Conv2d & $\left( 8, 32, 32 \right)$ & $(3, 1, 1)$ & ReLU \\
        Conv2d & $\left( 8, 32, 32 \right)$ & $(3, 1, 1)$ & ReLU \\
        MaxPool2d & $\left( 8, 16, 16 \right)$ & $(2, 0, 2)$ & None\\
        Dropout ($p=0.2$) & $8 \times 16 \times 16$ & None & None\\
        Linear & $c$ & None & Softmax\\
        \hline
        \end{tabular}
    \end{subtable}
    \vfill
    \vskip .1 in
    \begin{subtable}
        \centering
        \begin{tabular}{c|c|c}
        \multicolumn{3}{c}{FCN} \\
        \hline
        \rule{0pt}{2.0ex}
        Layer name & Output dim. & Activation function \\
        \hline
        \rule{0pt}{2.0ex}
        Input image & $3 \times 32 \times 32$ & None \\
        Linear & $d_l$ & ReLU \\
        Linear & $d_l$ & ReLU \\
        Linear & $c$ & Softmax\\
        \hline
        \end{tabular}
     \end{subtable}
     \label{table:CNN_FCN}
\end{table}

\vfill

\newpage

\subsection{Direct impact of label noise on minibatch gradients in SGD dynamics}

\begin{figure}[!ht]
    \vskip -0.1in
    \includegraphics{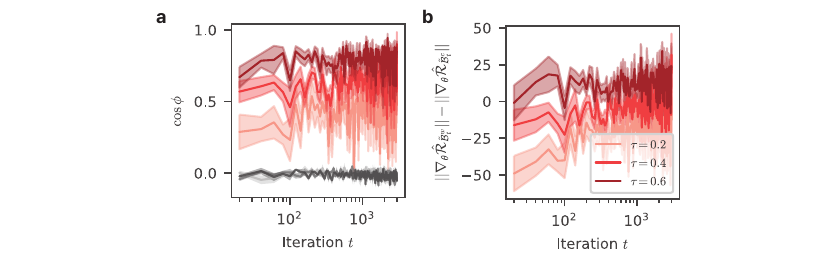}
    \vskip -0.1in
    \caption{(a) Cosine similarities between $-\bm{\nabla}_{\bm{\theta}} \mathcal{R}_{\tilde{\mathcal{B}}_{t}}(\bm{\theta})$ and $-\bm{\nabla}_{\bm{\theta}} \hat{\mathcal{R}}_{\tilde{\mathcal{B}}^w_{t}}(\bm{\theta}_t)$ ($\cos{\phi_{tw}}$; red), and between $-\bm{\nabla}_{\bm{\theta}} \mathcal{R}_{\tilde{\mathcal{B}}_{t}^c}(\bm{\theta})$ and $-\bm{\nabla}_{\bm{\theta}} \hat{\mathcal{R}}_{\tilde{\mathcal{B}}^w_{t}}(\bm{\theta}_t)$ ($\cos{\phi_{cw}}$; grey), respectively, throughout all training iterations for varying noise rate $\tau$.  
    (c) Magnitude difference between the two vectors $\| \bm{\nabla}_{\bm{\theta}}\hat{\mathcal{R}}_{\tilde{\mathcal{B}}_{t}^w}(\bm{\theta}) \| -\| \bm{\nabla}_{\bm{\theta}}\hat{\mathcal{R}}_{\tilde{\mathcal{B}}_{t}^c}(\bm{\theta}) \|$ throughout all training iterations for varying $\tau$. Here, we set the batch size to $B=8$ in setting $1$ described in Sec.~4. Darker colors represent larger values of $\tau$.
    }\label{supp3}
\end{figure}

In the previous sections, we demonstrated the impact of label noise on the drift in SGD dynamics. Here, we extend our analysis to examine the direct impact of label noise on the minibatch gradient. Using a similar decomposition as in Eq.~\eqref{app_eq:SGD_LE2}, the minibatch gradient can be expressed as:
\begin{equation}
\begin{aligned}
    \Delta \bm{\theta}_t &= -\frac{\eta}{B}\sum_{(\bm{x}_i, \bm{y}_i)\in \tilde{\mathcal{B}}_t} \bm{\nabla}_{\bm{\theta}} \mathcal{L}_i(\bm{\theta}_t) 
    \\ &= -\bm{\nabla}_{\bm{\theta}}{\mathcal{R}}_{\tilde{\mathcal{B}}_{t}}(\bm{\theta}_t) \eta = -\bm{\nabla}_{\bm{\theta}} \hat{\mathcal{R}}_{\tilde{\mathcal{B}}^c_{t}}(\bm{\theta}_t) \eta - \bm{\nabla}_{\bm{\theta}} \hat{\mathcal{R}}_{\tilde{\mathcal{B}}^w_{t}}(\bm{\theta}_t) \eta,
\end{aligned}
\label{app_eq:minibatch_gradients}
\end{equation}
where the corrupted minibatch $\tilde{\mathcal{B}}_t$ at the $t$-th iteration, sampled from $\tilde{\mathcal{D}}_{\rm tr}$, consists of $B^c_t$ correct labels and $B^w_t$ wrong labels (i.e., $B^c_t + B^w_t = B$). Here, $\bm{\nabla}_{\bm{\theta}} \hat{\mathcal{R}}_{\tilde{\mathcal{B}}^c_{t}}(\bm{\theta}_t) \equiv (B^c_t / B) \bm{\nabla}_{\bm{\theta}} {\mathcal{R}}_{\tilde{\mathcal{B}}^c_{t}}(\bm{\theta}_t)$ denotes the gradients from the correct part on minibatch, and $\bm{\nabla}_{\bm{\theta}} \hat{\mathcal{R}}_{\tilde{\mathcal{B}}^w_{t}}(\bm{\theta}_t) \equiv (B^w_t / B) \bm{\nabla}_{\bm{\theta}} {\mathcal{R}}_{\tilde{\mathcal{B}}^w_{t}}(\bm{\theta}_t)$ denotes the gradients from the wrong part on minibatch.
It is important to note that, unlike the decomposed drifts in Eq.~\eqref{app_eq:SGD_LE2} are determined for a given $\bm{\theta}$ and training dataset $\tilde{\mathcal{D}}_{\rm tr}$, the decomposed minibatch gradients in Eq.~\eqref{app_eq:minibatch_gradients} are inherently stochastic due to the randomness of the minibatch sampling process.
This stochasticity introduces greater fluctuations in the minibatch gradients compared to the drifts.
However, the fact that the average of minibatch gradients aligns with the corresponding drifts [e.g., $\langle \bm{\nabla}_{\bm{\theta}} \hat{\mathcal{R}}_{\tilde{\mathcal{B}}^w_{t}} (\bm{\theta}_t) \rangle = \bm{\nabla}_{\bm{\theta}} \hat{\mathcal{R}}_{\tilde{\mathcal{D}}^w_{\rm tr}}(\bm{\theta}_t)$] ensures consistent overall trends, i.e., the orthogonal relationship between $\bm{\nabla}_{\bm{\theta}} \hat{\mathcal{R}}_{\tilde{\mathcal{B}}^c_{t}}$ and $\bm{\nabla}_{\bm{\theta}} \hat{\mathcal{R}}_{\tilde{\mathcal{B}}^w_{t}}$ and the increasing contribution of $ \bm{\nabla}_{\bm{\theta}} \hat{\mathcal{R}}_{\tilde{\mathcal{B}}^w_{t}}$ with larger $\tau$, as shown in Fig.~\ref{supp3}.

\newpage

\subsection{Complementary plots for Secs.~4.1,~4.2,~4.3, and~4.4}
In order to facilitate a comparison between the resetting method and the original SGD, we normalized the validation loss and test accuracy values in the main text by calculating the relative difference in the metrics (RDVLoss and RDVAcc.). 
Here, we provide the unnormalized (i.e., original) validation loss and test accuracy values: Figs.~\ref{supp4},~\ref{supp5}, and~\ref{supp6} in the Supplementary Materials are the unnormalized results of Figs.~4, 5, and 6 in the main text, respectively. 
We additionally present test accuracies with and without resetting for varying noise rates in Fig.~\ref{fig:supp_high_ln} with the default setting of cross-entropy loss in Setting 2, and observe that resetting consistently improves performance even in the high-noise regime.

\begin{figure}[H]
    \centering
    \includegraphics[width=\linewidth]{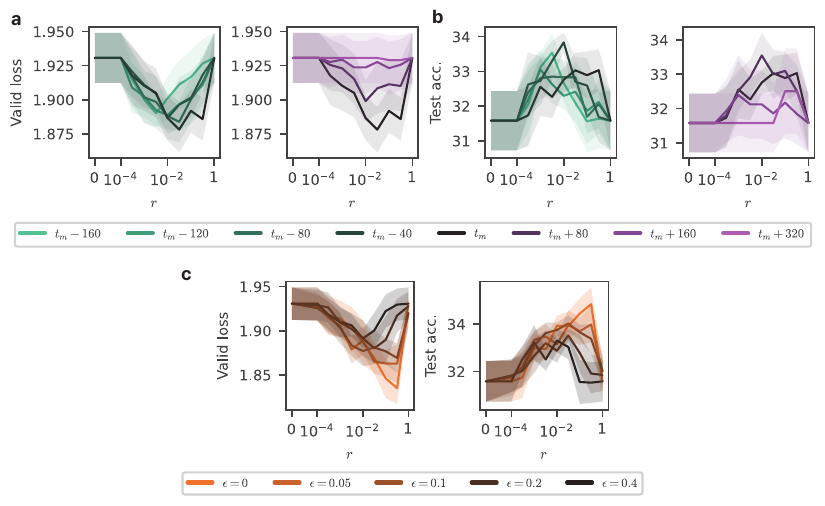}
    \vskip -0.1in
    \caption{(a) Validation loss and (b) test accuracy results with varying the checkpoint to reset to with respect to reset probability $r$.
    Based on the checkpoint at the overfitting iteration $t_{m}$, the results are obtained in earlier iterations (left) and later iterations than $t_m$ (right).
    Here, $t_m + \delta t$ denotes the iteration where the checkpoint is selected.
    (c) Validation loss (left) and test accuracy (right) with the perturbed checkpoint parameters $\bm{\theta}_{c, \epsilon}$. Here, $\bm{\theta}_{c, \epsilon} \equiv \bm{\theta}_c + \epsilon \hat{\bm{n}}$ where $\bm{\theta}_c$ denotes the checkpoint and $\hat{\bm{n}}$ denotes a random unit vector.
    The shaded areas denote the standard error.
    }\label{supp4}
    \vskip -0.1in
\end{figure}

\begin{figure}[H]
    \centering
    \includegraphics[width=\linewidth]{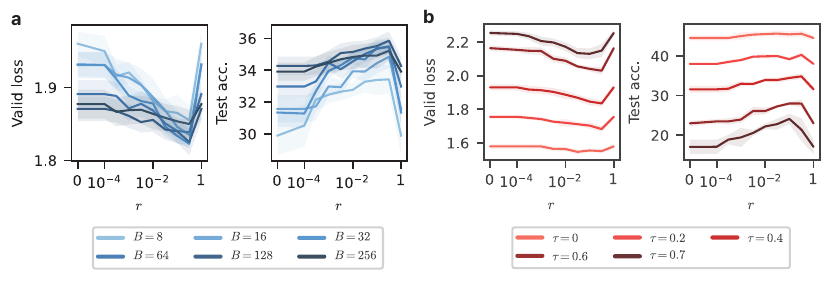}
    \vskip -0.1in
    \caption{Validation loss and test accuracy results with (a) varying the batch size $B$, and (b) varying the noise rate $\tau$ with respect to the reset probability $r$. We set $\tau = 0.4$ in (a) and $B=16$ in (b). The shaded areas denote the standard error.
    }\label{supp5}
    \vskip -0.1in
\end{figure}

\vfill

\begin{figure}[!t]
    \centering
    \includegraphics[width=\linewidth]{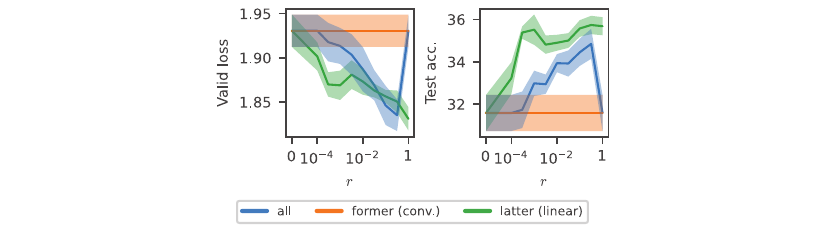}
    \vskip -0.1in
    \caption{Validation loss and test accuracy results with varying one section of the network to reset to with respect to reset probability $r$. Here, we set $\tau = 0.4$ and $B=16$. The shaded areas denote the standard error.
    }\label{supp6}
    \vskip -0.1in
\end{figure}

\begin{figure}[!t]
    \centering
    \includegraphics[width=\linewidth]{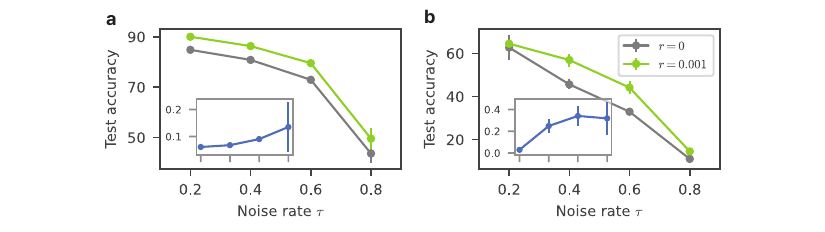}
    \vskip -0.1in
    \caption{Test accuracy results for (a) CIFAR-10 and (b) CIFAR-100 dataset by varying the noise rate $\tau$. Inset shows the relative difference in test accuracy (RDTAcc.) to represent the relative improvement compared to $r=0$.
    }\label{fig:supp_high_ln}
    \vskip -0.1in
\end{figure}

\subsection{Ablation study for ELR and SOP+}
\label{sec:ELR&SOP}
Since the hyperparameters $\lambda$ and $(\alpha_u, \alpha_v)$ for ELR and SOP+ baselines depend on the loss landscape (e.g., the dataset used, selection of model architecture, etc.), it is often costly to find optimal parameters in practical situations. To ensure that the stochastic resetting method can help users achieve high performance while not depending on the hyperparameter selection, we provide the test accuracy results for both optimal hyperparameter and non-optimal hyperparameter situations using the publicly available codes in Refs.~\cite{liu2020early-learning, liu2022robust}.
Table~\ref{table:elr-sop} shows the test accuracies for different settings of ELR and SOP+ baselines. Here, ELR and SOP+ denote the original methods with the hyperparameter sets used in Refs.~\cite{liu2020early-learning, liu2022robust}, while ELR* and SOP+* denote the methods with different hyperparameters. The hyperparameters for these methods are listed in Table~\ref{table:hyper-settings}.

The results corresponding to the use of optimal hyperparameters for each method (first and third rows in Table~\ref{table:elr-sop}, ELR and SOP+) show no significant improvement when stochastic resetting is used. On the other hand, under the non-optimal hyperparameter settings (the second and fourth rows, ELR* and SOP+*), the stochastic resetting method improves performance, especially for the CIFAR-100 dataset, which is more complex than CIFAR-10. These results suggest that stochastic resetting is compatible with existing powerful methods and can improve performance while also acting as a safeguard, regardless of the choice of hyperparameters for each method.

\begin{table}[!t]
\caption{Test accuracies (\%) on test datasets with ELR and SOP+ methods. We compare the performance without resetting (No) and with resetting (Reset) at $r=0.001$. Results are presented as the average and the standard deviation. The best results are indicated in \textbf{bold} with statistical significance.
}
\centering
\resizebox{\columnwidth}{!}{
\begin{tabular}{cccccccc}
\hline
\rule{0pt}{2.0ex}
\multirow{2}{*}{Dataset} & \multirow{2}{*}{Method} & \multicolumn{2}{c}{Noise rate $\tau = 0.2$} & \multicolumn{2}{c}{Noise rate $\tau = 0.4$} & \multicolumn{2}{c}{Noise rate $\tau = 0.6$}\\
& & No & Reset & No & Reset & No & Reset \\
\hline
\rule{0pt}{2.0ex}
    \multirow{4}{*}{CIFAR-10}   & ELR   & $91.9\pm0.2$ & $91.9\pm0.2$ & $90.2\pm0.2$ & $90.2\pm0.2$ & $87.3\pm0.3$ & $87.3\pm0.3$ \\
                                & ELR*  & $91.2\pm0.1$ & $\mathbf{91.3\pm0.2}$ & $88.8\pm0.2$ & $\mathbf{88.9\pm0.2}$ & $84.6\pm0.5$ & $84.6\pm0.5$ \\
                                & SOP+  & ${95.5\pm0.1}$ & $95.5\pm0.1$ & $94.8\pm0.2$ & $94.8\pm0.2$ & $\bm{93.6\pm0.2}$ & $93.5\pm0.2$ \\
                                & SOP+* & ${94.1\pm0.2}$ & $94.1\pm0.2$ & $89.9\pm0.3$ & $89.9\pm0.3$ & $85.0\pm0.3$ & $\mathbf{85.1\pm0.3}$ \\
\hline
\rule{0pt}{2.0ex}
    \multirow{4}{*}{CIFAR-100}  & ELR   & $\mathbf{73.2\pm0.2}$ & $73.0\pm0.4$ & $69.6\pm0.4$ & $69.6\pm0.4$ & $62.8\pm0.4$ & $62.8\pm0.4$ \\
                                & ELR*  & $67.4\pm0.3$ & $\mathbf{70.4\pm0.3^{\scriptscriptstyle ***}}$ & $55.1\pm0.6$ & $\mathbf{64.1\pm0.6^{\scriptscriptstyle ***}}$ & $45.9\pm0.7$ & $\mathbf{54.0\pm1.0^{\scriptscriptstyle ***}}$ \\
                                & SOP+  & ${78.4\pm0.2}$ & $\bm{78.5\pm0.1}$ & $76.8\pm0.3$ & $76.8\pm0.3$ & $73.6\pm0.6$ & $73.6\pm0.6$ \\
                                & SOP+* & ${72.1\pm0.3}$ & $72.1\pm0.3$ & $59.4\pm0.5$ & $\mathbf{65.7\pm0.9^{\scriptscriptstyle ***}}$ & $48.4\pm1.9$ & $\mathbf{54.8\pm2.3^{\scriptscriptstyle **}}$ \\
\hline
\end{tabular}}
\label{table:elr-sop}
\end{table}

\subsection{Test accuracies for datasets with asymmetric label noise}

Table~\ref{table:asymmetric} shows the test accuracy for the asymmetric noise case, while the rest of the settings remain the same as in Table~1. Similar to the results in Table~1, we find again that the stochastic resetting method makes robust improvement compared to no stochastic resetting. Similarly, results also show a similar trend in improvement as the noise rate increases, and more improvement is obtained in the CIFAR-100 dataset.

\begin{table}[!t]
\caption{Test accuracies (\%) on test datasets with asymmetric noise by different methods. We compare the performance without resetting (No) and with resetting (Reset) at $r=0.001$. Results are presented as the average and the standard deviation. The best results are indicated in \textbf{bold} with statistical significance.
}
\centering
\resizebox{\columnwidth}{!}{
\begin{tabular}{cccccccccc}
\hline
\rule{0pt}{2.0ex}
\multirow{2}{*}{Dataset} & \multirow{2}{*}{Method} & \multicolumn{2}{c}{Noise rate $0.1$} & \multicolumn{2}{c}{Noise rate $0.2$} & \multicolumn{2}{c}{Noise rate $0.3$} & \multicolumn{2}{c}{Noise rate $0.4$}\\
& & No & Reset & No & Reset & No & Reset & No & Reset \\
\hline
\rule{0pt}{2.0ex}
    \multirow{7}{*}{CIFAR-10}   & CE & $89.8\pm0.3$ & $\bm{93.3\pm0.3 ^{\scriptscriptstyle ***}}$ & $88.5\pm0.6$ & $\bm{91.6\pm0.6 ^{\scriptscriptstyle ***}}$ & $87.3\pm1.3$ & $\bm{90.0\pm0.8^{\scriptscriptstyle ***}}$ & $83.3\pm2.0$ & $\bm{86.7\pm0.8 ^{\scriptscriptstyle ***}}$\\
    & Part & --- & $\bm{92.6\pm0.3^{\scriptscriptstyle ***}}$ & --- & $\bm{90.9\pm0.4^{\scriptscriptstyle ***}}$ & --- & $\bm{89.6\pm0.6^{\scriptscriptstyle **}}$ & --- & $\bm{83.3\pm2.6}$ \\
    & MAE & ${89.1\pm4.2}$ & $89.1\pm4.2$ & $75.5\pm2.9$ & $\bm{75.6\pm3.0}$ & $57.0\pm0.1$ & $\bm{57.1\pm0.2}$ & $56.9\pm0.1$ & $\bm{56.9\pm0.1}$ \\
    & GCE & $92.2\pm0.1$ & $\bm{92.4\pm0.3}$ & $89.2\pm0.5$ & $\bm{89.9\pm0.2^{\scriptscriptstyle *}}$ & $85.3\pm0.7$ & $\bm{86.1\pm0.8^{\scriptscriptstyle ***}}$ & $78.0\pm1.6$ & $\bm{80.0\pm1.0^{\scriptscriptstyle **}}$\\
    & SL & $92.4\pm0.1$ & $\bm{92.5\pm0.2}$ & $90.1\pm0.1$ & $\bm{90.7\pm0.3^{\scriptscriptstyle *}}$ & $87.6\pm0.6$ & $\bm{87.9\pm0.3^{\scriptscriptstyle *}}$ & $81.1\pm0.9$ & $\bm{81.7\pm0.8^{\scriptscriptstyle ***}}$\\
    & ELR* & $93.4\pm0.2$ & $\bm{93.5\pm0.2}$ & $92.7\pm0.2$ & ${92.7\pm0.2}$ & $91.9\pm0.2$ & ${91.9\pm0.3}$ & ${90.4\pm0.2}$ & ${90.4\pm0.3}$ \\
    & SOP+* & $94.5\pm0.1$ & $\bm{94.6\pm0.2}$ & $94.0\pm0.1$ & $\bm{94.1\pm0.2}$ & $93.0\pm0.4$ & $93.0\pm0.4$ & $91.3\pm0.6$ & $\bm{91.5\pm0.2}$ \\
    \hline       
    \rule{0pt}{2.0ex}
    \multirow{7}{*}{CIFAR-100}   & CE & $\bm{72.0\pm0.5 ^{\scriptscriptstyle **}}$ & $69.3\pm0.9$ & $55.8\pm0.8$ & $\bm{65.5\pm1.5 ^{\scriptscriptstyle ***}}$ & $49.7\pm1.5$ & $\bm{58.6\pm1.7^{\scriptscriptstyle ***}}$ & $41.6\pm1.5$ & $\bm{49.0\pm1.6 ^{\scriptscriptstyle ***}}$\\
    & Part & --- & $\bm{69.2\pm0.9 ^{\scriptscriptstyle ***}}$ & --- & $\bm{65.2\pm1.1^{\scriptscriptstyle ***}}$ & --- & $\bm{58.8\pm0.5^{\scriptscriptstyle ***}}$ & --- & $\bm{48.7\pm1.3^{\scriptscriptstyle ***}}$\\ 
    & MAE & $21.7\pm2.4$ & $21.7\pm2.4$ & $17.5\pm1.7$ & $\bm{17.6\pm1.8}$ & $\bm{16.2\pm1.3}$ & $15.9\pm1.3$ & $14.2\pm2.1$ & $\bm{14.2\pm2.1}$ \\
    & GCE & $69.3\pm0.5$ & $\bm{70.0\pm0.2}$ & $62.0\pm0.8$ & $\bm{63.9\pm1.2^{\scriptscriptstyle **}}$ & $53.1\pm0.9$ & $\bm{55.2\pm1.5^{\scriptscriptstyle **}}$ & $41.7\pm0.5$ & $\bm{43.6\pm1.0^{\scriptscriptstyle *}}$\\
    & SL & $59.1\pm1.2$ & $\bm{66.0\pm1.2^{\scriptscriptstyle ***}}$ & $54.0\pm3.2$ & $\bm{60.1\pm2.7}$ & $50.0\pm1.1$ & $\bm{55.4\pm1.5^{\scriptscriptstyle *}}$ & $41.2\pm1.5$ & $\bm{46.1\pm1.0^{\scriptscriptstyle ***}}$\\
    & ELR* & $74.9\pm0.4$ & $74.9\pm0.4$ & $71.8\pm0.1$ & $71.8\pm0.1$ & $66.2\pm0.5$ & $\bm{68.4\pm1.6^{\scriptscriptstyle *}}$ & $57.2\pm0.6$ & $\bm{61.3\pm1.8^{\scriptscriptstyle **}}$\\
    & SOP+* & $74.3\pm0.4$ & $\bm{74.5\pm0.5}$ & $67.4\pm0.4$ & $\bm{70.9\pm0.5^{\scriptscriptstyle **}}$ & $59.6\pm0.5$ & $\bm{63.7\pm1.7^{\scriptscriptstyle **}}$ & $50.1\pm0.6$ & $\bm{52.9\pm1.9^{\scriptscriptstyle *}}$ \\
    \hline
\end{tabular}}
\label{table:asymmetric}
\end{table}

\subsection{Validation loss results in Sec.~4.4}

Table~\ref{table:LossTable} presents the validation loss results from the corresponding models used in Table~1 in the main text.
Similar to the test accuracy results in Table~1, Table~\ref{table:LossTable} shows that our resetting method consistently achieves either at least equivalent or lower validation losses compared to the baseline approach (i.e., no resetting). 

\begin{table}[H]
\caption{Validation losses with different methods. We compare the performance without resetting (No) and with resetting (Reset) at $r=0.001$. Results are presented as the average and the standard deviation. The best results are indicated in \textbf{bold} with statistical significance. 
}
\centering
\resizebox{\columnwidth}{!}{
\begin{tabular}{cccccccc}
\hline
\multirow{2}{*}{Dataset} & \multirow{2}{*}{Method} & \multicolumn{2}{c}{Noise rate $0.2$} & \multicolumn{2}{c}{Noise rate $0.4$} & \multicolumn{2}{c}{Noise rate $0.6$}\\
& & No & Reset & No & Reset & No & Reset \\
\hline
    \rule{0pt}{2.0ex}
    \multirow{7}{*}{CIFAR-10}   & CE & $1.231\pm0.018$ & $\bm{1.164\pm0.015^{\scriptscriptstyle ***}}$ & $1.774\pm0.010$ & $\bm{1.734\pm0.018^{\scriptscriptstyle **}}$ & $2.124\pm0.009$ & $\bm{2.102\pm0.012^{\scriptscriptstyle **}}$ \\
                                & PartRestart & --- & $\bm{1.184\pm0.012^{\scriptscriptstyle **}}$ & --- & $\bm{1.748\pm0.014^{\scriptscriptstyle *}}$ & --- & $\bm{2.114\pm0.011}$ \\
                                & MAE & $0.545\pm0.009$ & $0.545\pm0.009$ & $0.968\pm0.039$ & $\bm{0.963\pm0.039}$ & $1.422\pm0.042$ & $\bm{1.420\pm0.042}$ \\
                                & GCE & $0.388\pm0.008$ & $\bm{0.383\pm0.009}$ & $0.686\pm0.007$ & $\bm{0.673\pm0.007^{\scriptscriptstyle *}}$ & $0.957\pm0.009$ & $\bm{0.945\pm0.011}$ \\
                                & SL & $2.720\pm0.047$ & $\bm{2.707\pm0.041}$ & $4.759\pm0.045$ & $\bm{4.702\pm0.055}$ & $6.617\pm0.059$ & $\bm{6.558\pm0.069}$ \\
                                & ELR* & ${1.572\pm0.081}$ & ${1.572\pm0.081}$ & $\bm{1.718\pm0.021}$ & $1.723\pm0.022$ & $\bm{2.012\pm0.006}$ & ${2.016\pm0.006}$ \\
                                & SOP+* & ${2.067\pm0.035}$ & $\bm{2.029\pm0.061}$ & $\bm{1.611\pm0.007}$ & $1.632\pm0.021$ & $\bm{2.004\pm0.01}$ & ${2.011\pm0.012}$ \\

    \hline    
    \rule{0pt}{2.0ex}
    \multirow{7}{*}{CIFAR-100}  & CE & $2.759\pm0.053$ & $\bm{2.583\pm0.051^{\scriptscriptstyle ***}}$ & $3.675\pm0.064$ & $\bm{3.540\pm0.074^{\scriptscriptstyle *}}$ & $4.283\pm0.023$ & $\bm{4.201\pm0.026^{\scriptscriptstyle ***}}$ \\
                                & PartRestart & --- & $\bm{2.564\pm0.047^{\scriptscriptstyle ***}}$ & --- & $\bm{3.549\pm0.067^{\scriptscriptstyle *}}$ & --- & $\bm{4.206\pm0.019^{\scriptscriptstyle ***}}$ \\ 
                                & MAE & $1.690\pm0.039$ & $1.690\pm0.039$ & $1.864\pm0.044$ & $1.864\pm0.045$ & $1.932\pm0.010$ & $\bm{1.931\pm0.010}$ \\
                                & GCE & $0.644\pm0.007$ & $\bm{0.636\pm0.008}$ & $0.896\pm0.007$ & $\bm{0.883\pm0.006^{\scriptscriptstyle *}}$ & $1.129\pm0.004$ & $\bm{1.116\pm0.006^{\scriptscriptstyle **}}$ \\
                                & SL & $4.837\pm0.066$ & $\bm{4.727\pm0.066 ^{\scriptscriptstyle *}}$ & $6.578\pm0.036$ & $\bm{6.427\pm0.059^{\scriptscriptstyle **}}$ & $8.108\pm0.043$ & $\bm{7.996\pm0.038^{\scriptscriptstyle **}}$ \\
                                & ELR* & $3.393\pm0.072$ & $\bm{2.678\pm0.099^{\scriptscriptstyle ***}}$ & $3.985 \pm 0.477$ & $\bm{3.556\pm0.073}$ & $\bm{4.242\pm0.039}$ & ${4.253\pm0.068}$ \\
                                & SOP+* & $\bm{2.600\pm0.061}$ & ${2.605\pm0.047}$ & ${3.616\pm0.338}$ & $\bm{3.43\pm0.082}$ & ${4.185\pm0.074}$ & $\bm{4.180\pm0.088}$ \\
    \hline
\end{tabular}}
\label{table:LossTable}
\end{table}

\newpage

\subsection{Training time comparison \& scalability discussion for large-scale models}

\begin{figure}[!ht]
    \centering
    \includegraphics[width=\linewidth]{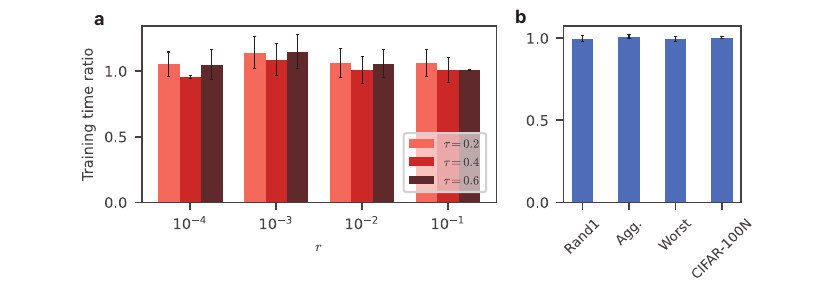}
    \vskip -0.1in
    \caption{
    Training time ratio of runs with resetting compared to runs with no reset (a) on ciFAIR-10 with varying the noise rate $\tau$ and the reset probability $r$ (Sec.~4.2), and (b) on real-world noisy datasets, CIFAR-10N/100N (Sec.~4.5). Error bars indicate the standard deviation.
    }\label{supp8}
    \vskip -0.1in
\end{figure}

In our experiments, we used DNN models up to ResNet-34, which has approximately 21M parameters and requires about 85MB in FP32 precision. 
For these models, the memory required to store the checkpoint did not significantly burden GPU memory, allowing us to maintain both the current model and the checkpoint in GPU memory during training.  
Consequently, this approach did not result in any noticeable increase in training time.
To validate this, we measure the training time ratios between runs with and without resetting in our experiments on the small dataset ciFAIR-10 (Setting 1 in Sec.~4.2) and the real-world noisy datasets CIFAR-10N/100N (Setting 3 in Sec.~4.5).
As shown in Fig.~\ref{supp8}, the training time ratios consistently fall within the error bars near 1 regardless of the reset probability $r$.
Note that the training time results for CIFAR-10N/100N in Setting 3 are almost the same as the results for CIFAR-10/100 in Setting 2.
These results demonstrate that our method does not involve significant time overhead when GPU memory is sufficient to maintain two DNN models simultaneously during training.

However, for much larger models, it may not be feasible to keep both the current model and the checkpoint in GPU memory.
In such cases, the checkpoint would need to be stored on a storage device and reloaded at each reset operation, potentially incurring additional I/O overhead. 
To estimate this cost for large-scale models, let us consider GPT-3~\cite{brown2020language}, which has 175B parameters (350GB in FP16), and single storage PCIe Gen4 x4 NVMe SSD with a transfer speed of 7GB/s.
Then, loading or saving the model would take approximately 50 seconds per operation.
GPT-3 was trained on 300B tokens with a global batch size of 3.2M tokens per iteration, resulting in approximately 94,000 iterations. 
Assuming 1,024 V100 GPUs~\cite{narayanan2021efficient}, the total training time for GPT-3 is calculated as 3.55 days.
Applying our method with $r = 10^{-3}$ to GPT-3 would only add 47,000 seconds (0.05 days) to the total training time, which is negligible in comparison to the overall training duration. 
Furthermore, while this calculation assumes the use of a single storage device, distributed storage systems are commonly employed for large-scale model training. These systems can significantly reduce loading times, minimizing the I/O overhead associated with resetting.

These findings demonstrate both the scalability and applicability of our method to large-scale models without significant time overhead. The performance improvements achieved on ViTs in Table 2 further support our claims. Additionally, the partial resetting method described in Sec.~4.3 offers an effective solution to mitigate I/O costs by resetting only portions of the DNN rather than the entire model, making it even more practical for large-scale implementations.

\vfill
\vfill
